\documentclass[sigconf]{acmart}

\usepackage{hyperref}
\usepackage{url}
\usepackage{booktabs} 
\usepackage[linesnumbered,ruled,vlined,algo2e]{algorithm2e}
\usepackage{multirow,enumitem}
\usepackage{subfig}
\usepackage{algorithm}
\usepackage{algorithmic}
\usepackage{siunitx}
\usepackage{listings}
\usepackage{amsthm}
\usepackage{xcolor}
\usepackage{balance}
\usepackage{arydshln}

\theoremstyle{definition}
\newtheorem{definition}{Definition}
\theoremstyle{plain}
\newtheorem{theorem}{Theorem}

\newtheorem{proposition}{Proposition}
\newtheorem{example}{Example}

\newcommand{\vpara}[1]{\vspace{0.05in}\noindent\textbf{#1 }}
\newcommand{\ours}{GraphAlign}

\newcommand{\RH}[1]{#1}
\newcommand{\XJ}[1]{#1}
\newcommand{\hide}[1]{}

\AtBeginDocument{%
  \providecommand\BibTeX{{%
    \normalfont B\kern-0.5em{\scshape i\kern-0.25em b}\kern-0.8em\TeX}}}

\copyrightyear{2024}
\acmYear{2024}
\setcopyright{acmlicensed}
\acmConference[KDD '24] {Proceedings of the 30th ACM SIGKDD Conference on Knowledge Discovery and Data Mining }{August 25--29, 2024}{Barcelona, Spain}
\acmBooktitle{Proceedings of the 30th ACM SIGKDD Conference on Knowledge Discovery and Data Mining (KDD '24), August 25--29, 2024, Barcelona, Spain}
\acmPrice{15.00}
\acmISBN{979-8-4007-0490-1/24/08}
\acmDOI{10.1145/3637528.3672058}
\begin{document}
\title[Can Modifying Data Address Graph Domain Adaptation?]{Can Modifying Data Address Graph Domain Adaptation?}

\author{Renhong Huang}

\affiliation{
  \institution{Zhejiang University}
  \city{Hangzhou}
  \country{China}
  }
\affiliation{
  \institution{Fudan University}
  \city{Shanghai}
  \country{China}
  }
\authornote{This work was done when the author was a visiting student at Fudan University.}
\email{renh2@zju.edu.cn}

\author{Jiarong Xu}
\affiliation{%
  \institution{Fudan University}
  \city{Shanghai}
  \country{China}
}
\authornote{Corresponding author.}
 \email{jiarongxu@fudan.edu.cn}
 
\author{Xin Jiang}
\affiliation{%
  \institution{Lehigh University}
  \city{Bethlehem}
  \country{United States}
}
\email{xjiang@lehigh.edu}

\author{Ruichuan An}
\affiliation{%
  \institution{Xi'an Jiaotong University}
  \city{Xi'an}
  \country{China}
}
\email{arctanx@stu.xjtu.edu.cn}

\author{Yang Yang}
\affiliation{%
  \institution{Zhejiang University}
  \city{Hangzhou}
  \country{China}
  }
\email{yangya@zju.edu.cn}

\renewcommand{\shortauthors}{Renhong Huang, Jiarong Xu, Xin Jiang, Ruichuan An, \& Yang Yang}


\begin{abstract}
Graph neural networks (GNNs) have demonstrated remarkable success in numerous graph analytical tasks. Yet, their effectiveness is often compromised in real-world scenarios due to distribution shifts, limiting their capacity for knowledge transfer across changing environments or domains. Recently, Unsupervised Graph Domain Adaptation (UGDA) has been introduced to resolve this issue. UGDA aims to facilitate knowledge transfer from a labeled source graph to an unlabeled target graph. Current UGDA efforts primarily focus on model-centric methods, such as employing domain invariant learning strategies and designing model architectures. However, our critical examination reveals the limitations inherent to these model-centric methods, while a data-centric method allowed to modify the source graph provably demonstrates considerable potential. This insight motivates us to explore UGDA from a data-centric perspective. By revisiting the theoretical generalization bound for UGDA, we identify two data-centric principles for UGDA: alignment principle and rescaling principle. Guided by these principles, we propose \ours, a novel UGDA method that generates a small yet transferable graph. By exclusively training a GNN on this new graph with classic Empirical Risk Minimization (ERM), \ours~attains exceptional performance on the target graph. Extensive experiments under various transfer scenarios demonstrate the \ours~outperforms the best baselines by an average of $2.16\%$, training on the generated graph as small as 0.25$\sim$1\% of the original training graph. 
\end{abstract}

\maketitle

\section{Introduction}
\vspace{-0.2in}
\begin{figure}[h]     
    \centering
    {\includegraphics[width=0.9\columnwidth]{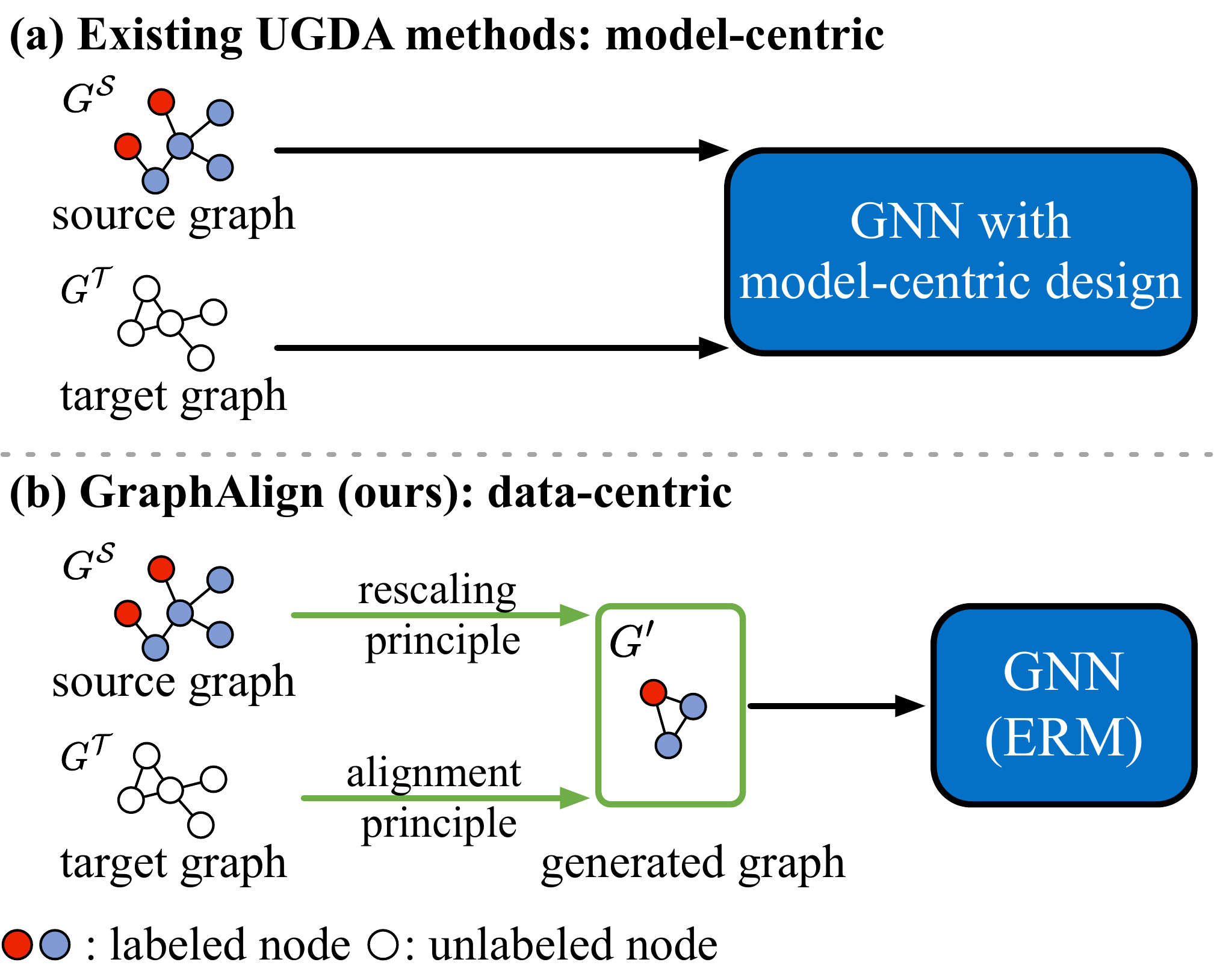}}
    \caption{Comparison between existing UGDA methods (which are all model-centric) and our data-centric method \ours. Guided by the rescaling and alignment principles, \ours~generates a small yet transferable graph, on which a simple GNN is trained with classic ERM. \ours~deviates from conventional approaches that employ sophisticated model design, and achieves outstanding practical performance.}\label{fig:intro}
\end{figure}
\vspace{-0.2in}

Graph is a ubiquitous data structure that models complex dependencies among entities. Typical examples include social networks~\cite{fan2019graph,davies2022realistic}, biological networks ~\cite{zhang2022ssgnn,ma2020multiview} and web networks~\cite{kleinberg1999web}.  Graph Neural Networks (GNNs) have demonstrated considerable potential in a variety of tasks related to graph data~\cite{sun2022beyond,huang2024measuring,fang2024universal,fang2024exploring,xu2022unsupervised,kuang2023unleashing,ma2023one}. However, they face notable challenges in real-world scenarios when distribution shift exists, \emph{e.g.}, GNNs are trained in one environment and then deployed in a different environment. This frequently occurs, for example, in social networks where interaction patterns among nodes change over time \cite{wang2021inductive}, and in molecular networks with diverse species \cite{cho2016compact}. 

To address distribution shift in GNNs, Unsupervised Graph Domain Adaptation (UGDA) has emerged as a crucial solution. The goal of UGDA is to take advantage of the labeled source graph to facilitate the transfer of knowledge to an unlabeled target graph. Most existing efforts in UGDA have been made from the model-centric perspective, \textit{e.g.}, employing domain invariant learning strategies and designing model architectures, as depicted in Figure~\ref{fig:intro} (a). Specifically, they aim to learn consistent representations across different domains by minimizing domain discrepancy metrics~\cite{Shen_2021,zhu2021shift,you2023graph} or by incorporating adversarial training with a domain discriminator~\cite{wu2022eerm,dai_graph_2022,wu2020UDAGCN}.

In this work, we investigate the inherent limitations of existing model-centric UGDA methods. We identify scenarios where, regardless of how the model parameters are modified, these methods consistently fail in node classification tasks. On the contrary, we find that a data-centric approach that is allowed to modify source graph can theoretically achieve an arbitrarily low classification error with classic empirical risk minimization (ERM) setup. This highlights the potential of data-centric methods in addressing UGDA. 

Inspired by these findings, we aim to tackle UGDA from a data-centric perspective. By revisiting the theoretical generalization bound for UGDA~\cite{shen2018wasserstein}, we propose two data-centric principles: (1) \textit{Alignment principle} suggests reducing the discrepancy between the modified source graph and the target graph. (2) \textit{Rescaling principle} states that a smaller graph modified from the source graph can achieve a generalization error comparable to that of a larger graph.

With the guidance of these principles, we present a data-centric UGDA method, aiming to generate a new graph that is significantly smaller than the original source graph, yet retains enough information from the source graph and effectively aligns with the target graph. The purpose of such a data-centric UGDA method is to achieve outstanding performance on the target graph with a GNN trained on the generated graph with standard ERM but without sophisticated model design, as shown in Figure~\ref{fig:intro} (b).  In essence, we encounter three main difficulties: (1) how to measure and compute the distribution discrepancy of non-Euclidean structure of graphs when aligning the generated graph with the target graph; (2) how to ensure that the generated graph (with a much smaller size) retains sufficient information from the source graph; and (3) how to efficiently optimize the generated graph, which involves multiple interdependent variables related to graph structure, node features, and node labels.

To address the above difficulties, we propose a novel UGDA method, named \ours. We first use a surrogate GNN model to map graphs into Euclidean representation spaces, and utilize a computationally more efficient Maximum Mean Discrepancy (MMD) distance as the discrepancy metric. The generated graph is expected to have a much smaller size (inspired by rescaling principle) and at the same time, to be so informative that a GNN model trained on the generated graph behaves similarly to that trained on the source graph. To achieve this, we match the gradients of the GNN parameters w.r.t.\ the generated graph and the source graph. Then, during the optimization process, we model the graph structure as a function of node features, so the main decision variables are only the node features. We also introduce a novel initialization approach for the generated graph, inspired by the theoretical connection between GNN transferability and the spectral distance to the target graph. This initialization is shown to enhance practical performance and accelerate the optimization process.

Our contributions are summarized as follows:
\vspace{-0.1in}
\begin{itemize}[leftmargin = 10pt]
    \item \textbf{New perspective}: For the first time, UGDA is addressed from a data-centric perspective.
    
    \item \textbf{New principles}:  Inspired by a theoretical generalization bound for UGDA, we propose two data-centric principles that serve as the guidelines for modifying graph data: the alignment principle and the rescaling principle.
    
    \item \textbf{New method}: We propose \ours, a novel UGDA method. \ours~generates a small yet transferable new graph, on which a simple GNN model is trained using classic ERM. In particular, \ours~does not need sophisticated GNN design and achieves outstanding performance on target graphs.
    
    \item \textbf{Extensive experiments}: Experiments on four scenarios and twelve transfer setups demonstrate the effectiveness and efficiency of our method in tackling UGDA. In particular, our method beats the best baselines by an average of +2.16\% and trains only on a smaller generated graph, with size up to 1\% of the original training graph.
\end{itemize}
The rest of the paper is organized as follows.  We first present the paradigm of UGDA and basic definitions in \S\ref{sec:pre}. Then, we present the limitations of existing UGDA methods and propose data-centric principles for UGDA in \S\ref{sec:method}. Guided by these principles, we show in \S\ref{sec:model} that the limitation of existing UGDA methods can be mitigated by using our proposed model \ours, which employs a data-centric approach that generates a small yet transferable graph for GNN training. Finally, we evaluate the effectiveness and efficiency of our method in \S\ref{sec:exp}.
\vspace{-0.1in}
\section{Preliminaries} \label{sec:pre}
In this section, we present the basic paradigm of UGDA, and introduce the contextual stochastic block model (CSBM), which will be used in \S\ref{sec:method} to build a motivating example.

\vpara{Unsupervised Graph Domain Adaptation (UGDA).}
We focus on UGDA for node classification tasks, where we have a labeled source domain graph and an unlabeled target domain graph. We denote the labeled source graph as $G^{\mathcal{S}} = \big(A^{\mathcal{S}}, X^{\mathcal{S}}, Y^{\mathcal{S}} \big)$ with {$n^\mathcal S$} nodes, where $A^{\mathcal{S}}$, $X^{\mathcal{S}}$ and $Y^{\mathcal{S}}$ represent the adjacency matrix, node features, and node labels of the source graph, respectively. 
The unlabeled target graph is denoted by $G^\mathcal T = \big( A^\mathcal T, X^\mathcal T \big)$ with {$n^\mathcal T$} nodes, where $A^\mathcal T$ and $X^\mathcal T$ are the adjacency matrix and node features, respectively. 

Given the labeled source graph $G^{\mathcal{S}}$ and unlabeled target graph~$G^{\mathcal{T}}$, UGDA aims to train a GNN $h$ that predicts accurately the node labels {$Y^\mathcal T$} of target graph. The GNN $h = g \circ f$  typically consists of a feature extractor $f: \mathcal G \rightarrow \mathcal Z$ and a classifier $g: \mathcal Z \rightarrow \mathcal Y$, where $\mathcal G$, $\mathcal Z$ and $\mathcal Y$ represent input space, representation space, and label space, respectively. A common approach of UGDA is to learn invariant representations by ensuring that the feature extractor $f$ outputs representations whose distribution remains consistent across both source and target graphs \cite{wu2020UDAGCN,zhu2021shift,dai_graph_2022}.

\vpara{Contextual Stochastic Block Model.}
The contextual stochastic block model (CSBM) is an integration of the stochastic block model (SBM) with node features for random graph generation~\cite{deshpande2018contextual}. CSBM generates a graph based on a prescribed edge connection probability matrix, and the distribution for node features is determined by distinct Gaussian mixture models for each class. In the context of UGDA, two distinct CSBMs can be used to model the source and target graphs, facilitating the examination of the domain shift. Before we formally define CSBM, we remark that CSBM is only used to build the motivating example in \S\ref{sec:method} and is \textit{not} needed in the design of our model. For simplicity, we consider the CSBM with two classes.

\begin{definition}[Contextual Stochastic Block Model]
CSBM is a generative model that builds a labeled graph $G = (A, X, Y)$ (with node size {$n$}) as follows. The node labels  are random variables drawn from a Bernoulli distribution (\emph{e.g.},  {$Y_i \sim \operatorname{Bern} (0.5)$}). The entries of the adjacency matrix follow a Bernoulli distribution {$a_{ij} \sim \operatorname{Bern} (C_{pq})$}  if node $i$ belongs to class $p$ and $j$ belongs to class $q$, where the matrix $C \in [0,1]^{2 \times 2}$ is a prescribed  probability matrix that is used to model edge connections. Node features are drawn independently from normal distributions $X_i \sim \mathcal N(\boldsymbol{\mu}, I)$ if $Y_i = 0$ and $X_i \sim \mathcal N(\boldsymbol{\nu}, I)$ if $Y_i = 1$, where $I$ is the identity matrix. Such a CSBM is denoted by $\operatorname{CSBM} ({n}, C, {\boldsymbol{\mu}, \boldsymbol{\nu}})$.
\end{definition}
\vspace{-0.1in}
\section{Data-Centric Principles} \label{sec:method}

In \S\ref{subsec:example}, we present a constructive example to demonstrate the inherent limitation in current model-centric UGDA methods that only focus on sophisticated GNN model design. This example further motivates our exploration of data-centric principles for UGDA in \S\ref{subsec:bound}. 

\vspace{-0.1in}
\subsection{Motivating Example} \label{subsec:example}
To understand the potential issues in existing UGDA models, we investigate the performance of GNN models on a constructive graph pair $(G^\mathcal S, G^\mathcal T)$.  Example~\ref{example} describes the source and target graphs constructed via CSBMs. Proposition~\ref{prop:model} identifies the limitation of existing model centric-based UGDA approaches, that are typically designed with the goal of learning domain-invariant representations. Proposition \ref{prop:data} shows the potential benefits of adopting a data-centric approach in UGDA.

\begin{example} \label{example}
Consider the source and target graphs generated by two CSBMs: $\operatorname{CSBM}(n, C^{\mathcal{S}},\boldsymbol{\mu},\boldsymbol{\nu})$ and $\operatorname{CSBM} (n, C^{\mathcal{T}}, \boldsymbol{\tilde{\mu}}, \boldsymbol{\tilde{\nu}})$, respectively. In both CSBMs, each class consists of $n/2$ nodes, and their edge connection probability matrices are
\[
C^{\mathcal{S}} = \left[\begin{array}{cc}
    a & a \\
    a & a-\Delta
\end{array}\right], \qquad C^{\mathcal{T}} = \left[\begin{array}{cc}
    a-\Delta & a \\
    a & a
\end{array}\right],
\]
where $a$ and $\Delta$ are constants with $0 < \Delta < a < 1$. 
\end{example}

The following proposition shows that existing UGDA model-centric methods focusing solely on sophisticated GNN model design would fail even for the simple case in Example~\ref{example}.

\begin{proposition} \label{prop:model}
Assuming the feature extractor $f$ is a single-layer GNN, and it is trained with the domain-invariant constraint $\mathbb{P}(f(G^{\mathcal{S}})) \\ = \mathbb{P}(f(G^{\mathcal{T}}))$, and then used for inference on the target graph. When such a GNN $f$ is applied to Example~\ref{example}, the classification error in the target domain is always larger than a strictly positive constant, regardless of the parameters of the GNN.
\end{proposition}
The proofs of Proposition  \ref{prop:model} can be found in Appendix~\ref{app:proofs}. Proposition~\ref{prop:model} suggests that model-centric UGDA models would fail the node classification task for the graphs in Example~\ref{example}. In comparison, if we are allowed to ``modify'' the source graph, it could yield a GNN model with an arbitrarily small classification error, as shown in the following proposition.

\begin{proposition} \label{prop:data}
Suppose that the feature extractor $f$ is a single-layer GNN. Also, suppose that a data-centric approach is employed to construct a new graph $G^\prime$ by modifying $G^\mathcal S$ with the constraint $\mathbb P(G^\prime)  = \mathbb P(G^{\mathcal{T}})$. The GNN $f$ is trained with standard ERM on $G^\prime$, which minimizes the classification error on $G^\prime$, and then used for inference on the target graph. There exist examples of graphs generated in Example~\ref{example} such that the classification error in the target domain is arbitrarily small.
\end{proposition}

The proofs of Proposition~\ref{prop:data} can be found in Appendix~\ref{app:proofs}. Proposition~\ref{prop:data} highlights that, in certain cases, adapting the data can be more beneficial than adapting the model. An intuitive reason is that data-centric approaches that modify the source data could mitigate the inherent difference of the data distribution between the source and target domains  (\emph{i.e.}, $\mathbb P(G^\prime) = \mathbb P(G^\mathcal T)$). Thus, the GNN model trained on the modified graph $G^\prime$ can be seamlessly applied to the target graph, resulting in a reduction in the error in the target domain. Overall, Propositions~\ref{prop:model} and~\ref{prop:data} reveal the potential benefits of a data-centric approach for UGDA.

\vspace{-0.1in}
\subsection{Data-Centric Principles for UGDA}\label{subsec:bound}

As demonstrated previously, our objective is to address UGDA through a data-centric approach. Before we delve into the detailed methods, we first discuss two principles that serve as the guidelines for modifying source data. Specifically, we present a generalization bound for UGDA that   lays the theoretical foundation for these principles. In our case, the generalization error is defined as the classification error in the target domain \cite{shen2018wasserstein}:
\[
    \epsilon^{\mathcal{T}}(g,f)= \mathbb{E}_{\mathbb P(G^{\mathcal{T}})} \big( \| g \circ f(G^{\mathcal{T}}) -\psi^{\mathcal{T}}(G^{\mathcal{T}}) \| \big),
\]
where $\psi^\mathcal T: \mathcal G^\mathcal T \to \mathcal Y^\mathcal T$ is the true labeling function on the target graph. Based on~\cite{shen2018wasserstein}, we can derive the generalization bound in the following theorem.

\begin{theorem}[Generalization bound for UGDA \cite{shen2018wasserstein}] \label{thm:genbound}
Denote by $L_\mathrm{GNN}$ the Lipschitz constant of the GNN model $g \circ f$. Let the hypothesis set be $\mathcal{H} = \left\{h = g \circ f : \mathcal G \rightarrow \mathcal Y \right\}$, and let the pseudo-dimension be $\operatorname{Pdim}(\mathcal{H}) = d$. The following inequality holds with a probability of at least $1-\delta$:
\begin{equation}
\begin{aligned}
\epsilon^{\mathcal{T}}(g, f) &\leq \hat{\epsilon}^{\mathcal{S}}(g, f) + \eta
+ \underbrace{2 L_{\mathrm{GNN}} W_1\big(
{\mathbb P(G^{\mathcal{S}})}, \mathbb P(G^{\mathcal{T}})\big)}_{\text{alignment term}} \\
&\quad \mathrel{+} \underbrace{\sqrt{\frac{4 d}{{n^{\mathcal{S}}}} \log \left(\frac{e {n^{\mathcal{S}}}}{d}\right)+\frac{1}{
{n^{\mathcal{S}}}} \log \left(\frac{1}{\delta}\right)}}_{\text{rescaling term}}, 
\end{aligned} \label{eq:bound}
\end{equation}
where $\hat \epsilon^{\mathcal S} (g,f) = ({1}/{n^\mathcal S}) \| g \circ f(G^{\mathcal{S}}) -\psi^{\mathcal{S}}(G^{\mathcal{S}}) \|$ is the empirical classification error in source domain with $\psi^\mathcal S$ the true labeling function on the source domain, $\eta = \min_{h \in \mathcal H} \big\{\epsilon^{\mathcal{S}}(g^{*}, f^{*}) + \epsilon^{\mathcal{T}}(g^{*}, f^{*})\big\}$ denotes the optimal combined error that can be achieved on both source and target graphs by the optimal hypothesis $g^{*}$ and $f^{*}$, $\mathbb P(G^{\mathcal{S}})$ and $\mathbb P(G^{\mathcal{T}})$ are the graph distribution of source and target domain respectively, the probability distribution $\mathbb P(G)$ of a graph $G$ is defined as the distribution of all the ego-graphs of $G$, and $W_1(\cdot, \cdot)$ is the Wasserstein distance. 
\end{theorem}

\vspace{-0.1in}

Proof of Theorem~\ref{thm:genbound} can be found in Appendix~\ref{app:proofs}. The last two terms on the right-hand side of~\eqref{eq:bound}, labeled as the alignment term and rescaling term, inspire the following two principles that will serve as critical guidelines when modifying source data.

\vpara{Alignment principle:}
\emph{\XJ{Modifying the source graph to minimize the distribution discrepancy between the modified source and target graphs can reduce the generalization error.}}

This principle is inspired by the alignment term in the bound~\eqref{eq:bound}. The smaller the divergence of distribution between the source and target graphs (\emph{i.e.}, $W_1(\mathbb P(G^{\mathcal{S}}), \mathbb P(G^{\mathcal{T}}))$), the smaller the generalization bound.

\begin{figure}[t]     
    \centering
    \includegraphics[width=0.7\columnwidth]{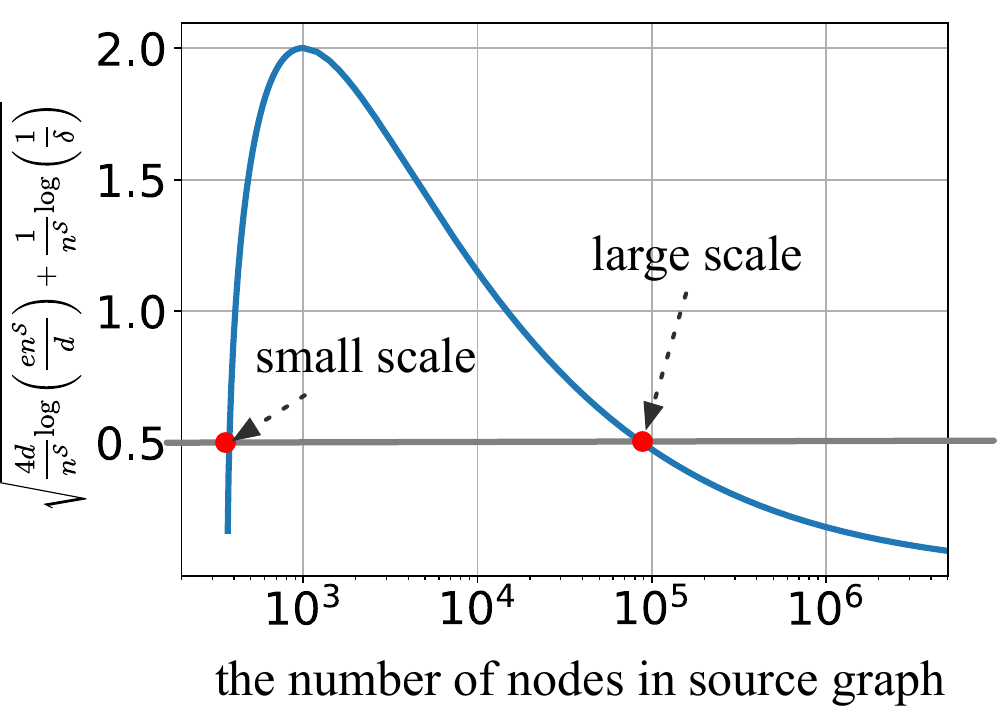}
    \vspace{-0.05in}
    \caption{The figure illustrates how the rescaling term varies with the scale of the source graph. We specify $\delta=0.01$ to ensure that the \eqref{eq:bound} holds with a probability of at least 99\%. The pseudo-dimension $d$ is set to $1000$, which is a reasonable assumption based on~\cite{devroye1996vapnik} (note that the trend of the rescaling term's variation is consistent, regardless of the value of $d$). The horizontal axis is presented on a logarithmic scale.}
    \label{fig:function}
\end{figure}

\vpara{Rescaling principle:}  
\emph{\XJ{Modifying the source graph to reduce its scale could achieve a generalization error comparable to that of a larger-scale source graph.}}

This principle encourages us to decrease graph size to improve efficiency while achieving comparable accuracy. It is drawn from the behavior of the rescaling term in the bound~\eqref{eq:bound}. Figure~\ref{fig:function} shows that this term increases at first, and then decreases as the node size further grows. Notably, a smaller-scale source graph is capable of achieving an accuracy comparable to that of a larger-scale source graph (indicated by the red points in the figure). Yet, as expected, an overly small node size is not advisable, as it loses too much information.

We further discuss the first two terms in the bound~\eqref{eq:bound}. The first term is related to the performance of the source domain, and it is a common practice to enhance source domain performance in order to obtain a good performance on the target domain~\cite{shen2018wasserstein,redko2017theoretical,wu2020UDAGCN,dai_graph_2022}. Regarding the second term, it is often overlooked by other unsupervised domain adaptation methods~\cite{you2023graph,wu2020UDAGCN,zhu2021shift,dai_graph_2022,shen2018wasserstein,redko2017theoretical}. Given that previous research~\cite{liu2019transferable} have demonstrated that generating transferable examples by confusing domain discriminator can effectively bridge the domain divergence and reduce this term, similar effects can be achieved by modifying the source graph to better align with the target graph by our approach. 
\vspace{-0.1in}
\section{Proposed Method: \ours}\label{sec:model} 
In this section, we propose a novel UGDA method \ours. \ours \ strictly adheres to the alignment and rescaling principles and generates a new graph to replace the source graph for training. These two principles guide us in the generation of a new graph that (1) is much smaller than the original source graph, (2) aligns well with the target graph, and (3) retains enough information from the source graph.

\begin{definition}[Data-Centric UGDA]~\label{problem:data_ugda}
Given the labeled source graph $G^{\mathcal{S}} = \big(A^{\mathcal{S}}, X^{\mathcal{S}}, Y^{\mathcal{S}} \big)$ with $A^{\mathcal{S}} \in \mathbb R ^ {n^\mathcal{S} \times n^\mathcal{S}}, X^{\mathcal{S}} \in \mathbb R ^ {n^\mathcal{S} \times d}, Y^{\mathcal{S}} \in \{0, \cdots, c-1 \}^{n^\mathcal{S}}$,  and the unlabeled target graph $G^\mathcal T = \big( A^\mathcal T, X^\mathcal T \big)$, Data-Centric UGDA generates a new graph $G^\prime=(A^\prime, X^\prime, Y^\prime)$ with $A^\prime \in \mathbb R ^ {n^\prime \times n^\prime}, X^\prime \in \mathbb R ^ {n^\prime \times d}, Y^\prime \in \{0, \cdots, c-1 \}^{n^\prime}$  and $n^\prime \ll n^\mathcal{S}$. The graph $G^\prime$ is designed to (1) align with the target graph $G^{\mathcal{T}}$, and (2) incorporate sufficient information from the source graph $G^{\mathcal{S}}$,  such that the GNN model trained with standard ERM on $G^\prime$ rather than $G^{\mathcal{S}}$ yields enhanced performance on $G^{\mathcal{T}}$.
\end{definition}

Next, we will introduce the problem for optimizing $G^\prime$ in \S\ref{subsec:obj}. Following this, we describe our approach to relaxing the optimization problem and modeling the generated graph $G^\prime$ in \S\ref{subsec:rescale}. We finally provide the complexity analysis of our method in \S\ref{subsec:complexity}.

\vspace{-0.1in}
\subsection{Optimization Problem} \label{subsec:obj}
We here formulate the optimization problem that guides the alignment of graph~$G^\prime$ with the target graph while incorporating sufficient information from the source graph.

\vpara{Enhancing generalization based on alignment principle.}
The alignment principle tells us to achieve a lower  generalization bound, and it is better to align the distribution of the generated graph $G^{\prime}$ more closely with that of the target graph $G^{\mathcal{T}}$.  Referring to the alignment term in the generalization bound, this can be achieved by optimizing $G^{\prime}$ so as to minimize the Wasserstein distance:
\vspace{-0.1in}
\begin{equation} \label{eq:wasserstein}
W_1(\mathbb P(G^\prime), \mathbb P(G^{\mathcal{T}})) = \inf _{\gamma \in \Gamma(\mathbb P(G^\prime), \mathbb P(G^{\mathcal{T}}))} \mathbb{E}_{(u, v) \sim \gamma} \ c(u, v),
\end{equation}

where $u$ and $v$ are ego-graphs sampled from $\mathbb P(G^\prime)$ and $\mathbb P(G^\mathcal{T})$ respectively,  $c(u,v)$ is the distance function between the ego-graphs $u$ and $v$, $\Gamma$ is the set of all joint distribution of $\gamma \in \Gamma(\mathbb P(G^\prime), \mathbb P(G^{\mathcal{T}}))$, and the marginals for $\gamma$ are $\mathbb P(G^\prime)$ and $\mathbb P(G^{\mathcal{T}})$ on the first and second factors respectively.

Although the Wasserstein distance is a natural objective in the optimization problem, its minimization remains a great challenge for the following reasons. First, calculating the Wasserstein distance for a given pair of $(u,v)$ involves solving a large-scale linear program, which is itself computationally expensive, let alone the minimization of the Wasserstein distance. What's worse, the computation and minimization of the Wasserstein distance in the non-Euclidean space, such as graph data, are even more difficult. We present our solutions as follows.

First, the computational complexity of the Wasserstein distance grows cubicly in the problem dimension~\cite{pele2009fast}, which is unacceptably expensive in our case. To resolve this, a common alternative to the Wasserstein distance is the MMD distance, which is computationally cheaper and more efficient~\cite{bikowski2018demystifying,  birdal2020synchronizing}. 

However, replacing Wasserstein distance with the MMD distance does not resolve the second issue caused by non-Euclidean structure of graphs. To handle this issue, existing efforts often use graph kernels that map graphs into (Euclidean) representation spaces~\cite{nikolentzos2017matching,chuang2022tree,titouan2019optimal}. Yet, such mapping process is typically non-differentiable, which complicates the optimization process, and computing graph kernels is still unacceptably costly in our case. In this work, we employ a surrogate GNN model to represent the mapping process. A good example of the GNN model is GIN \cite{xu2019powerful}, owing to its discriminative power akin to the Weisfeiler-Lehman test~\cite{weisfeiler1968reduction}.

Consequently, we can update generated graph $G^\prime$ by minimizing the following objective derived from the alignment principle as
\begin{equation}
\mathcal{L}_{\mathrm{alignment}}= \mathrm{MMD}\left(\hat{\mathbb P}(\text{GNN}(A^{\prime}, X^{\prime})), \hat{\mathbb P}(\text{GNN}(A^{\mathcal{T}}, X^{\mathcal{T}}))\right),
\end{equation}
where $\hat{\mathbb P}$ is the empirical distribution computed via random sampling. Details regarding the implementation details for MMD distance are provided in Appendix~\ref{app:exp-setup}.

\vpara{Incorporating sufficient information from source graph.} 
The generated graph $G^\prime$ should retain enough information from the source graph, which can be guaranteed by the following two strategies.

First, if $G^\prime$ is adequately informative as the source graph $G^{\mathcal{S}}$, a GNN model trained on $G^\prime$ would behave similarly to that trained on $G^{\mathcal{S}}$. Inspired by~\cite{zhao2020dataset}, we aim to match the gradients of the GNN parameters w.r.t. $G^{\prime}$ and $G^{\mathcal{S}}$. This is crucial for preserving the essential information from the source graph in $G^\prime$. To achieve this, we focus on minimizing the following objective function:
\[
    \resizebox{1.0\hsize}{!}{$ 
\mathcal{L}_{\mathrm{mimic}}=\operatorname{Cos} \left(\nabla \mathcal{L}_{\mathrm{CE}} \left(\text{GNN}(A^{\prime},X^{\prime}), Y^{\prime}\right), \nabla \mathcal{L}_{\mathrm{CE}}(\text{GNN}(A^{\mathcal{S}},X^{\mathcal{S}}), Y^{\mathcal{S}}) \right)
$},
\]
where $\mathcal{L}_{\mathrm{CE}}$ denotes the cross entropy loss, GNN is the surrogate GNN model and $\operatorname{Cos}$ is the cosine similarity function. 

On the other hand, the generated graph $G^\prime$ needs to reflect generally observed properties in real-world networks. A typical property of real-world networks is feature smoothness, where connected nodes often share similar features~\cite{mcpherson2001birds,belkin2001laplacian}. Moreover, real-world graphs are usually sparse~\cite{zhou2013learning}.  Therefore, to ensure that $G^\prime$ accurately represents these real-world characteristics, we focus on minimizing the following objective function:
\begin{equation} \label{eq:prop}
\mathcal{L}_{\mathrm{prop}}=\operatorname{tr}(X^{\prime T}LX^{\prime})+\|A^{\prime}\|_F^2,
\end{equation}
where $L=I-D^{-\frac{1}{2}}A^{\prime}D^{-\frac{1}{2}}$ is the normalized Laplacian matrix and $D$ is the diagonal degree matrix for $A^{\prime}$. The first term in $\mathcal L_\mathrm{prop}$ captures feature smoothness while the second one characterizes sparsity.

\vpara{Optimization problem.}
We here outline the construction of the generated new graph $G^{\prime}$. Specifically, our goal is to build $G^\prime = (A^\prime, X^\prime, Y^\prime) \in \mathbb R^{n^\prime \times n^\prime} \times \mathbb R^{n^\prime \times d} \times \mathbb R^{n^\prime}$ so that it aligns with the target graph and retains enough information from the source graph. Integrating the aforementioned objectives, the construction of $G^\prime$ can be formulated as the following optimization problem 
\begin{equation} \label{eq:opt-1}
 \underset{A^{\prime}, \, X^{\prime}, \, Y^\prime}{\min} \;\; \mathcal{L}_{\mathrm{mimic}}+ \alpha_1 \mathcal{L}_{\mathrm{alignment}}+ \alpha_2 \mathcal{L}_{\mathrm{prop}},
\end{equation}
where $\alpha_1, \alpha_2 >0$ are hyper-parameters.

\vspace{-0.1in}
\subsection{Modeling the Generated Graph} \label{subsec:rescale}
Note that the decision variables in \eqref{eq:opt-1} are $A^\prime$, $X^\prime$, and $Y^\prime$. Optimizing the three variables is extremely difficult due to their interdependence. To this end, the node size $n^\prime$ is pre-chosen and proportional to $n^{\mathcal S}$, \textit{i.e.}, $n^\prime = r n^{\mathrm S}$ for some prescribed $0 < r \ll 1$. We also require that the node labels $Y^\prime$ have the same distribution as $Y^{\mathcal S}$ when randomly choose $n^\prime$ nodes from the source graph.


Even if $n^\prime$ is pre-fixed and much smaller than $n$, the number of parameters in $A^\prime$ is still quadratic in $n^\prime$ and prohibitively large to optimize. To further reduce the number of parameters in $A^\prime$, we propose to model the graph structure $A^{\prime}$ as a function of $X^{\prime}$. This is motivated by the observation in real-world networks that the graph structure and the node features are implicitly correlated~\cite{pfeiffer2014attributed}. So, in our implementation, $A^\prime$ is modeled as
\[
\resizebox{1.0\hsize}{!}{$ 
A^{\prime} = \rho_{\phi}(X^{\prime}), \text{with} \ A^{\prime}_{ij}
\sim \operatorname{Bernoulli}\left(\operatorname{Sigmoid} \left(\operatorname{MLP}_{\phi}\left(X^{\prime}_i, X^{\prime}_j\right)\right)\right),
$}
\]
where $\rho_{\phi}$, parameterized by $\phi$, is the function that transforms node features to graph structure, and $\operatorname{MLP}_{\phi}$ is a multi-layer neural network. As is common in the literature, the non-differentiability of Bernoulli sampling can be handled by the Gumbel-Max reparametrization technique~\cite{jang2017categorical}.

In summary, the decision variables in \eqref{eq:opt-1} are effectively reduced to $X^\prime$ and $\phi$, and then the original problem~\eqref{eq:opt-1} can be rewritten as
\begin{equation}~\label{eq:summary}
 \underset{ X^{\prime}, \, \phi}{\min} \;\;
 \mathcal{L}_{\mathrm{mimic}}+ \alpha_1 \mathcal{L}_{\mathrm{alignment}}+\alpha_2 \mathcal{L}_{\mathrm{prop}}.
\end{equation}

\vspace{-0.1in}
\vpara{Initialization of the generated graph.}
As one may expect, the initial value of $X^\prime$ is crucial for solving~\eqref{eq:summary}. A good initialization not only helps improve practical performance, but can also accelerate the optimization process.

To further improve the performance of proposed method, we propose an initialization strategy for $X^\prime$. Intuitively, we hope to select from the source graph those nodes and features that already present transferability. This intuition can be further supported by the following theorem, which builds theoretical connections between the property of the newly generated $G^\prime$ and the transferability of GNNs.

\begin{theorem}[GNN transferability] \label{thm:transferability}
Let $G^\prime$ and  $G^{\mathcal{T}}$ be the newly generated graph and the target graph. Given a GNN graph encoder~$f$, the transferability of the GNN $f$ satisfies
\begin{equation}
\left\|f(G^{\prime})-f(G^{\mathcal{T}})\right\|_2
\leq \xi_1
\Delta_\mathrm{spectral}\left(G^{\prime}, G^{\mathcal{T}}\right)+\xi_2,
\end{equation}
where $\xi_1$ and $\xi_2$ are two positive constants, and $\Delta_\mathrm{spectral}\left(G^{\prime}, G^{\mathcal{T}}\right) = \tfrac{1}{n^{\prime}n^{\mathcal{T}}} \sum_{i=1}^{n^{\prime}} \sum_{j=1}^{n^{\mathcal{T}}} \|L_{G^\prime_i}-L_{G^{\mathcal T}_j} \|_2$ measures the spectral distance between $G^{\prime}$ and $G^{\mathcal{T}}$. Here $G^\prime_i$ is the ego-graph of node $i$ in $G^{\prime}$, and $L_{G^\prime_i}$ is its normalized graph Laplacian. The graph Laplacian $L_{G^{\mathcal T}_j}$ is defined in a similar manner.
\end{theorem}

Theorem~\ref{thm:transferability} suggests that a smaller spectral distance between $G^{\prime}$ and $G^{\mathcal{T}}$ indicates better transferability. Based on this interpretation, we propose to select $n^\prime$ nodes from $G^{\prime}$ whose features are used as the initial values of $X^\prime$. Such selection guarantees that the graph $G^\prime$ constructed by these nodes and features has a small spectral distance $\Delta_\text{spectral}(G^{\prime}, G^{\mathcal{T}})$. 

\vspace{-0.1in}
\subsection{Complexity Analysis} \label{subsec:complexity}
The traditional domain adaptation methods typically involve GNN training and domain-invariant learning, \emph{e.g.}, using MMD loss for optimization. The time complexity of these methods is typically $O(d^2n+dn^{2})$.  In comparison, the time complexity of \ours~primarily depends on the optimization process of $G^{\prime}$. Assume the dimension of representations is denoted as $d$ and the number of nodes in $G^{\prime}$ is $n^{\prime}$. Constructing $G^{\prime}$ involves calculating ${n^{\prime}}^2$ edges, which is equivalently $O\left( r^{2}dn^{2} \right)$. Regarding the GNN forward pass, it requires a time complexity of $O\left( d^{2}n \right)$. For the computation of $\mathcal{L}_\mathrm{alignment}$, the time complexity is $O\left( rdn^{2} \right)$ because of the computation of kernel function. For the $\mathcal{L}_\mathrm{mimic}$ loss, we need to inference GNN on both $G^{\mathcal{S}}$ and $G^{\prime}$, resulting in a time complexity of $O\left( d^{2}n+rd^{2}n \right)$. Besides, the time complexity of computing $\mathcal{L}_{\mathrm{prop}}$ primarily focuses on the calculations of $\operatorname{trace}(X^{\prime T}LX^{\prime})$ an is $O\left(r^{2}n^{2}\right)$, with the trace operation requires $O\left( rn \right)$. Overall, the time complexity of \ours~is determined by $O( (r^2d+rd+r^2)n^2+$ \\ $(1+r)d^2n)$. In our experiments, the hyper-parameter $r$ is chosen as $r=0.01$, so the complexity of \ours~is much cheaper compared with classic model-centric UGDA methods.
\vspace{-0.1in}
\section{Experiments} \label{sec:exp}
In this section, we evaluate the performance of \ours~under various transfer scenarios. We first generate a new graph $G^\prime$ using the proposed method, and then train a GNN on $G^\prime$ with classic ERM. The GNN is then tested on the target graph. The experiments span a range of transfer scenarios, and we also include ablation studies, hyper-parameter analysis, and runtime comparison to demonstrate the effectiveness of \ours. 

\vspace{-0.15in}
\subsection{Experimental Setup}~\label{subsec:setup}
\vspace{-0.2in}

\vpara{Datasets.} We conduct experiments on node classification in the transfer setting across six scenarios. In each scenario, we train the GNN on one graph and evaluate it on the others.
\begin{itemize}[leftmargin=1em]
\item{\textbf{ACMv9 (A), DBLPv7 (D), Citationv1 (C)}~\cite{dai_graph_2022}}: These datasets are citation networks from different sources, where each node represents a research article and an edge indicates citation relationship between two articles. The data are collected from ACM (prior to 2008), DBLP (between 2004 and 2008), and Microsoft Academic Graph (after 2010), respectively. We include six transfer settings: C$\rightarrow$D, A$\rightarrow$D, D$\rightarrow$C,  A$\rightarrow$C,  D$\rightarrow$A and C$\rightarrow$A.

\item \textbf{{$\text{ACM}_\text{small}$ ($\hat{\text{A}}$), $\text{DBLP}_\text{small}$ ($\hat{\text{D}}$)}}~\cite{wu2020UDAGCN}: These two are also citation networks, with articles collected between the years 2000 and 2010, and after year 2010. We include two transfer settings: {$\hat{\text{A}} \rightarrow \hat{\text{D}}$, $ \hat{\text{D}} \rightarrow \hat{\text{A}}$}.

\item\textbf{Cora-degree, Cora-word}~\cite{gui2022good}: They are two transfer settings for citation networks provided by \cite{bojchevski2018deep}, derived from the full Cora dataset~\cite{bojchevski2018deep}. The data pre-process involves partitioning the original Cora dataset into two graphs based on node degrees and selected word count in publications, respectively. Each setting evaluates the transferability from one graph to the other.

\item \textbf{Arxiv-degree, Arxiv-time}~\cite{gui2022good}: They are two transfer settings for citation networks provided by \cite{bojchevski2018deep}, adapted from the Arxiv dataset that comprises computer science arXiv papers~\cite{bojchevski2018deep}. The partitioning of Arxiv into two graphs is based on node degrees and time, respectively. 

\item \textbf{USA, Brazil, Europe}~\cite{
ribeiro2017struc2vec}: They are collected from transportation statistics and primarily comprise airline activity data, where each node represents to an airport. We include six transfer settings: USA$\rightarrow$Brazil, USA$\rightarrow$Europe, Brazil$\rightarrow$USA, Brazil$\rightarrow$Europe, Europe$\rightarrow$USA, Europe$\rightarrow$Brazil.

\item \textbf{Blog1, Blog2}~\cite{
shen2020adversarial}: They are collected from the BlogCatalog dataset, where node represents a blogger, and edge indicates friendship between bloggers. The node attributes comprise keywords extracted from self-descriptions of blogger, and the task is to predict their corresponding group affiliations.  We include two transfer settings:  Blog1$\rightarrow$Blog2, Blog2$\rightarrow$Blog1.
\end{itemize}
\vspace{-0.1in}

\vpara{Baselines.}
We compare our method with the following baselines, which can be categorized into three classes:
(1) \textbf{Vanilla ERM}, including GCN~\cite{kipf2017semisupervised}, GraphSAGE ~\cite{hamilton2017inductive} and GIN~\cite{xu2019powerful}. They are trained on the source graph with ERM and then directly evaluated on the target graph. (2) \textbf{Non-graph domain adaptation methods}, including MMD~\cite{long2015learning}, CMD~\cite{Zellinger_2019}, DANN~\cite{ajakan2014domain}, CDAN~\cite{long2018conditional} are considered.  To adapt them to the UGDA setting,  we replace the encoder with GCN~\cite{kipf2017semisupervised}. (3) \textbf{UGDA methods}, including UDAGCN~\cite{wu2020UDAGCN}, AdaGCN~\cite{dai_graph_2022}, MIXUP~\cite{han2022gmixup}, EERM~\cite{wu2022eerm}, MFRReg~\cite{you2023graph}, SSReg~\cite{you2023graph}, GRADE~\cite{wu2023non}, JHGDA~\cite{shi2023improving} and STRURW~\cite{liu2023structural}. 

\vpara{Implementation details.}
We evaluate our proposed method by training a GCN~\cite{kipf2017semisupervised} on the generated graph $G^\prime$ with ERM and test the GCN on $G^\mathcal T$. When computing $\mathcal{L}_{\mathrm{mimic}}$, we train a surrogate GCN on $G^\prime$ for the supervised node classification task, with a cross-entropy loss. When computing $\mathcal{L}_{\mathrm{alignment}}$, we train GIN on $G^{\prime}$ under the infomax principle following~\cite{velivckovic2018deep}. All the GNN models adopt a two-layer structure with 256 hidden units, while the other hyper-parameters are set to default.  After training all the surrogate models, we freeze all the parameters when optimizing $G^{\prime}$.  We set the reduction rate $r=0.25\%$ for Arxiv (due to its large scale) and $r=1\%$ for the remaining datasets. The values of $\alpha_1$, $\alpha_2$ are set to 1 and 30. For the optimizer, we use Adam~\cite{kingma2017adam} with a learning rate of \num{1e-3} and weight decay of \num{5e-3}. We use mini-batch training with batch size 32. The total iterations of training is 300. 

When evaluating the baselines, for Vanilla ERM and non-graph domain adaptation baselines, we employ two-layer GCN with 256 hidden units and the remaining hyper-parameters are set to default values. We follow the setting of ~\cite{wu2020UDAGCN} to perform a {grid} search on the trade-off between classification loss and the loss function designed to address domain adaptation, exploring values within [0.01, 0.1, 1.0, 10.0] and reporting the best performance. Adam is employed for optimization with a learning rate of \num{1e-3} and weight decay of \num{5e-3}. We use mini-batches of size 32 over 300 training iterations. For UGDA methods, we adopt their default hyper-parameters.

The reported numbers in all experiments are the mean and standard deviation over 10 trials. More details can be found in Appendix~\ref{app:exp-setup}.
Our codes are available at \url{https://github.com/zjunet/GraphAlign}.

\begin{table*}[t]
\caption{Micro F1 scores across various transfer settings on citation networks. The {bold} numbers denote the best result. ``OOM'' denotes the instances where the method ran out of memory.}
  \vspace{-0.1in}
    \resizebox{2.1\columnwidth}{!}{
    \centering
    \setlength \tabcolsep{1.2pt}
    \renewcommand{\arraystretch}{1.3}
\begin{tabular}{c|cccccc|cc|cc|cc|c} 
 \toprule[1.5pt]
 & \multicolumn{6}{c}{\small\textbf{ACMv9 (A), DBLPv7 (D), Citationv1 (C)}}&\multicolumn{2}{c}{\small\textbf{$\text{ACM}_\text{small}$ ($\hat{\text{A}}$), $\text{DBLP}_\text{small}$ ($\hat{\text{D}}$)}} &\multicolumn{2}{c}{\small\textbf{Cora}}&\multicolumn{2}{c}{\small\textbf{Arxiv}}\\ 
Methods   & C$\rightarrow$D    & A$\rightarrow$D    & D$\rightarrow$C  & A$\rightarrow$C   & D$\rightarrow$A   & C$\rightarrow$A  &$\hat{\text{A}} \rightarrow \hat{\text{D}}$ &$\hat{\text{D}} \rightarrow \hat{\text{A}}$ &Cora-word & Cora-degree  & Arxiv-degree & Arxiv-time  &Avg.rank \\ \hline
ERM~(GIN) & 43.32(2.40) &39.32(2.10) &38.86(1.57)&37.27(2.92)&37.14(1.36)&35.40(2.71)& 50.12(3.06)& 63.24(1.53)& 30.64(1.63)& 19.91(1.76)&22.69(1.51)&  32.25(2.42) & 14.9\\ 
ERM~(SAGE) &64.22(0.89)&61.73(0.88)&60.92(1.25)&61.90(1.19)&55.66(0.92)&57.66(1.04)&77.51(3.17)& 36.59(4.05)& 61.63(0.18)& 53.56(0.30)&47.84(0.52)& 44.84(0.56) &8.3\\
ERM~(GCN)       &67.80(3.49)&61.05(0.37)&62.36(5.01)&61.78(4.33)&53.78(4.13)&62.93(5.32)& 63.28(2.05)&  68.48(0.89)&63.26(0.35)& 54.42(0.51)&43.86(1.12)&  12.86(5.16) & 7.8\\  \hline
DANN & 66.02(1.89)&61.44(2.89)&54.68(3.66)&59.61(4.88)&49.01(3.59)&55.02(2.45)&63.38(2.21)&68.53(0.90)&63.24(0.41)&  54.44(0.54)& 43.86(1.12)& 12.87(5.17)  & 9.8 \\ 
CDAN &53.69(3.90)&61.53(1.64)&61.13(2.78)&	60.48(2.61)&53.69(3.90)&	58.22(0.95) &63.53(2.09)&70.73(0.86)&63.42(0.27)&  54.48(0.42)&43.91(1.00)&  12.86(5.18) & 8.2\\
MMD  &63.06(0.61)&59.72(0.35)&59.46(0.49)&62.98(0.50)&53.57(0.33)&59.34(0.47)& 61.20(0.69)&69.22(0.87)&63.27(0.38)&  54.21(0.60)&OOM & OOM & 10.25\\
CMD &47.89(12.49)&46.67(11.03)&48.49(5.13)&53.02(10.79)&44.8(2.83)&49.31(8.70)& 50.56(6.17)&64.86(3.73)&54.95(0.64)&  49.61(0.48)&39.13(1.40)&  13.58(0.85) &13.3  \\ \hline
UDAGCN&70.70(2.64)&64.64(3.12)&56.34(8.55)&62.40(5.81)&49.57(4.95)&55.92(5.85)&69.97(4.10)&70.43(1.36)&63.40(0.21)&  53.98(0.41)&37.82(6.70)&47.44(0.77)&7.3 \\
AdaGCN& 69.72(2.05)&67.67(0.92)&66.38(2.86)&69.34(1.44)&56.78(2.53)& 63.34(1.24)& 69.28(2.34)&69.33(1.57)& 62.91(0.49) & 53.24(0.47) &38.93(1.62) & 12.38(5.50)&5.8\\
MIXUP   &67.60(1.88)&63.08(2.68)&60.75(5.95)&66.04(3.36)&54.01(5.00)&62.06(2.35)&51.10(1.40)&68.43(1.46)&65.44(5.95)& 62.79(4.01)&52.13(0.40)& 23.63(0.24)&6.3\\ 
EERM  &43.60(1.59)&46.27(5.36)&43.97(2.74)&46.37(3.25)&38.39(1.89)&43.87(2.10)&67.05(1.33)&47.11(4.82)&13.10(0.77)& 7.42(0.86)&OOM&OOM&14.9\\ 
GRADE& 64.68(1.30) & 54.46(1.13) & 59.61(0.09) & 66.05(0.35) & 57.30(0.18) & 61.19(0.46) & 64.92(0.09) & 55.47(0.21) & 48.37(2.45)& 42.84(1.87)&	41.36(1.64)&11.32(0.78)& 9.8\\ 
JHGDA& 65.09(4.98) & 58.20(1.33) & 51.31(3.56) & 64.51(2.65)&	46.46(4.72)	&59.25(3.44) & 71.51(2.40) & 62.26(3.28) & OOM &OOM &OOM &OOM & 12.3\\ 
MFRReg &66.99(7.77)&61.39(8.45)&65.71(6.33)&70.72(8.71)&56.65(7.21)&59.81(8.53)&65.80(10.42)&71.13(0.95)& 59.23(1.81)& 53.04(1.28)&OOM&OOM&8.3\\ 
SSReg  &63.36(17.73)&61.78(9.98)&66.53(4.47)&61.91(11.87)&56.05(9.41)&60.41(6.75)&71.10(8.31)&70.00(1.41)&59.61(2.26)&  52.19(1.29)&OOM&OOM&9.1\\
STRURW& 64.10(0.35)	&59.45(0.60)&58.91(1.02)&63.42(0.38)&55.83(1.11)&	62.41(1.27)&77.47(1.01)& 72.11(2.21)&62.41(0.72)& 67.76(0.39)& 57.45(0.15)&49.98(0.12)&5.9\\ 
\hline
\ours&\textbf{72.56(0.61)}&\textbf{69.65(0.26)}&\textbf{68.08(0.32)}&\textbf{75.61(0.24)}&\textbf{62.06(0.68)}&\textbf{67.36(0.40)}&\textbf{79.51(3.75)}&\textbf{72.63(2.21)}&\textbf{66.37(1.46)}&\textbf{69.83(1.53)}& \textbf{57.51(1.42)} & \textbf{51.17(1.37)} &\textbf{1.0} \\
\bottomrule
\end{tabular}
}\label{table:results}
\end{table*}

\begin{table*}[!h]
\caption{Micro F1 scores across various transfer settings on airport networks and social networks.
  }\label{table:other-results}
    \vspace{-0.1in}
    \resizebox{1.7\columnwidth}{!}{
    \centering
    \setlength \tabcolsep{1.2pt}
    \renewcommand{\arraystretch}{1.3}
    \begin{tabular}{c|cccccc|cc|c}
 \toprule[1.5pt]
~ & \multicolumn{6}{c}{\small\textbf{Airport}} & \multicolumn{2}{c}{\small\textbf{Social}}  \\ 
Methods  & USA$\rightarrow$Brazil & USA$\rightarrow$Europe & Brazil$\rightarrow$USA & Brazil$\rightarrow$Europe & Europe$\rightarrow$USA & Europe$\rightarrow$Brazil & Blog1$\rightarrow$Blog2 & Blog2$\rightarrow$Blog1  &Avg.rank \\ \hline
ERM (GIN) & 35.88(4.71) & 32.33(1.78) & 41.43(9.32) & 33.18(5.72) & 44.52(5.72) & 42.90(7.39) & 18.36(0.25) & 19.47(1.14) & 15.5 \\ 
ERM (SAGE) & 49.62(1.37) & 43.96(0.61) & 53.29(0.40) & 53.78(1.01) & 52.79(0.68) & 67.63(1.04) & 45.40(0.61) & 47.00(1.05) &  6.9\\ 
ERM (GCN) & 43.36(4.81) & 37.99(4.39) & 49.95(0.82) & 42.21(1.87) & 56.82(0.34) & 71.76(0.84) & 44.57(1.24) & 41.25(2.34)&8.9  \\ \hline
DANN & 59.85(8.34) & 52.48(2.09) & 53.38(0.33) & 57.74(1.61) & \textbf{57.48(0.48)} & 70.99(0.68) & 42.30(1.23) & 41.30(2.67)& 4.8 \\ 
CDAN & 46.87(3.82) & 42.61(1.79) & 52.29(0.99) & 45.01(1.51) & 56.76(0.39) & 72.06(1.57) & 42.56(1.68) & 41.21(3.32)&8.0 \\ 
MMD & 52.90(0.78) & \textbf{55.64(0.69)} & 52.49(0.33) & 56.41(0.31) & 56.77(0.14) & 72.98(0.37) & 43.98(1.83) & 40.98(2.26)&  4.6  \\ 
CMD & 60.84(0.57) & 54.99(0.88) & 48.99(0.70) & 58.50(0.75) & 55.21(0.68) & 72.98(1.50) & 28.65(5.31) & 26.55(6.42)& 7.3  \\ \hline
UDAGCN & 34.42(3.14) & 51.78(1.06) & 24.96(6.12) & 44.81(1.93) & 55.45(0.44) & 44.43(2.60) & 36.47(7.45) & 36.89(5.48)&12.8  \\ 
AdaGCN & 52.37(3.79) & 48.67(0.58) & 45.63(4.44) & 51.48(2.61) & 48.97(4.23) & 66.11(3.82) & 38.25(1.86) & 36.82(3.65)&11.5   \\ 
MIXUP & 43.18(4.13) & 41.63(0.66) & 50.34(4.73) & 42.74(0.98) & 49.32(1.62) & 68.76(1.48) & 40.31(1.84) & 39.63(2.82) &11.6 \\ 
EERM & 39.12(5.87) & 43.42(0.90) & 42.98(6.40) & 55.72(1.91) & 48.92(2.20) & 48.92(6.37) & 41.89(2.67) & 40.21(1.96)& 11.5  \\
GRADE& 24.43(2.44)&	24.81(4.33)&	34.29(1.42)&27.32(1.90)&34.03(2.35)&29.77(3.53)&17.33(1.17)&16.39(1.42)& 16.9\\ 
JHGDA&61.41(4.73)&52.09(3.73)&	 51.42(5.71)&58.04(9.05)&51.88(2.94)&72.77(7.43)&17.86(2.56)& 17.93(2.60)& 8.4 \\ 
MFRReg & 56.99(6.16) & 53.19(6.45) & 50.71(5.07) & 50.72(8.71) & 56.65(7.21) & 69.18(8.53) & 40.84(3.12) & \textbf{46.34(6.72)} & 7.1\\ 
SSReg & 53.36(7.73) & 53.78(1.72) & 49.34(4.88) & 52.43(6.23) & 54.28(6.22) & 55.71(6.25) & 40.93(4.29) & 45.37(8.11)& 8.3 \\ 
STRURW & 60.73(0.34) & 53.77(0.98) & 52.19(2.01) & 53.48(0.23) & 49.67(2.88) & 63.40(1.27) & 46.02(0.95) & 38.64(1.76)& 7.5  \\ \hline
\ours & \textbf{62.90(0.78)} & 54.32(0.94) & \textbf{54.38(0.10)} & \textbf{58.80(0.86)} & 57.34(2.02) & \textbf{73.12(0.90)} & \textbf{47.14(1.72)} & 45.83(5.01)&\textbf{1.6} \\ \hline
    \end{tabular}
}
\end{table*}

\subsection{Experimental Results}\label{subsec:cross-domain}
\vspace{-0.1in}
\vpara{Main results.}
Table~\ref{table:results} and Table~\ref{table:other-results} presents the experiment results across various settings. For all the datasets, \ours~surpasses all baselines and shows an average improvement of +2.16\% over the best baseline. Notably, this superior performance is attained with our generated small graphs. In comparison, the suboptimal performance of Vanilla ERM and non-graph domain adaptation methods highlights the critical need for a carefully tailored UGDA strategy to address domain discrepancy on graphs. Compared with other UGDA methods, our superior performance emphasizes the advantage of adopting a data-centric approach over a model-centric approach in UGDA.

\begin{figure}[t]     
    \centering
    \includegraphics[width=1.0\columnwidth]{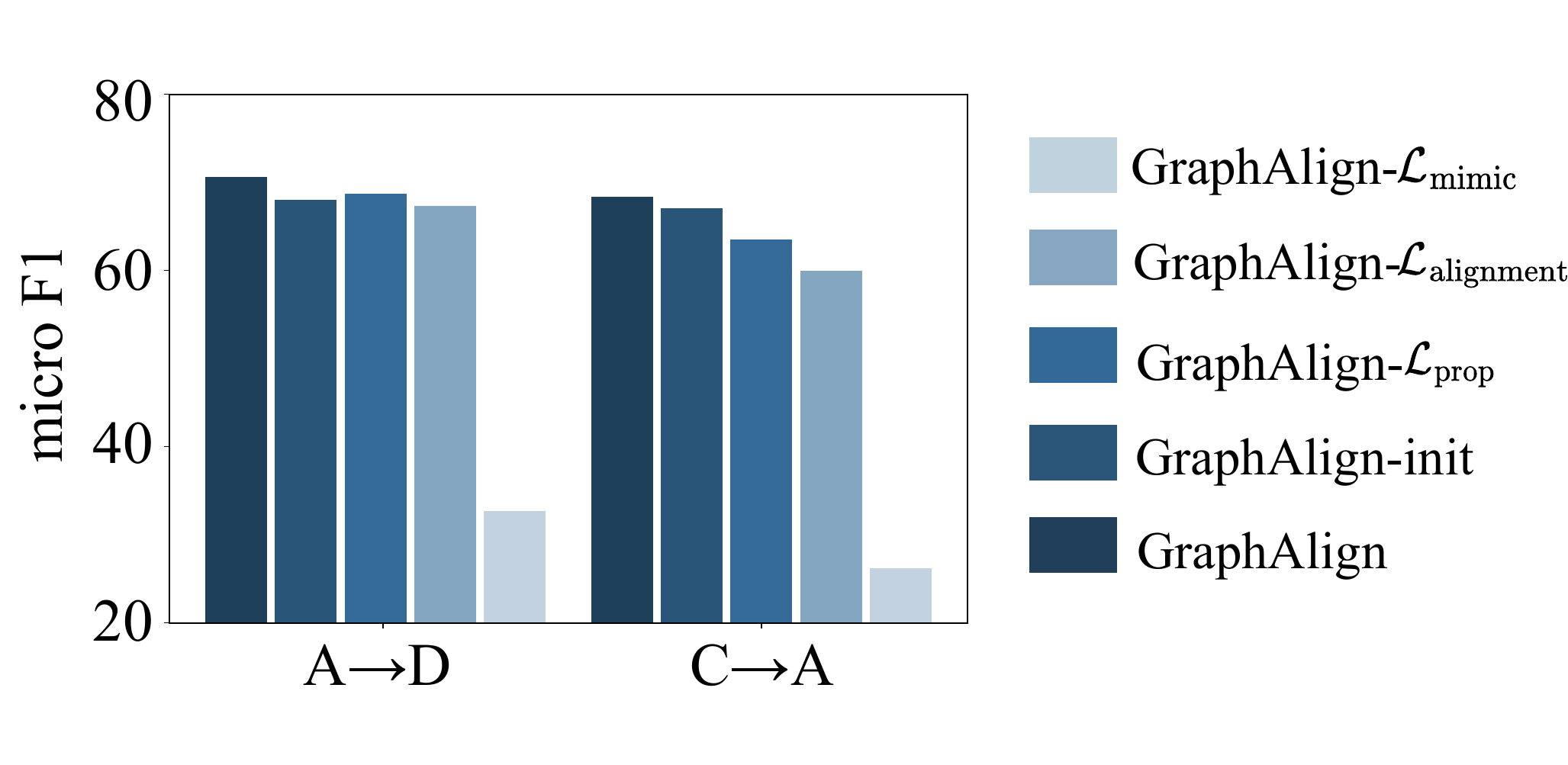}
    \vspace{-0.3in}
    \caption{Ablation studies on A$\rightarrow$D and C$\rightarrow$A tasks.}
    \label{fig:ablation}
    \vspace{-0.1in}
\end{figure}

\vpara{Ablation studies.}
To validate the effectiveness of components, ablation studies are conducted on: (1) \ours-init, employing random initialization instead of our proposed initialization;  (2) \ours-$\mathcal{L}_{\mathrm{prop}}$, removing the loss $\mathcal{L}_{\mathrm{prop}}$. (3) \ours-$\mathcal{L}_{\mathrm{alignment}}$, removing the loss $\mathcal{L}_{\mathrm{alignment}}$. (4) \ours-$\mathcal{L}_{\mathrm{mimic}}$, removing the loss $\mathcal{L}_{\mathrm{mimic}}$. The results, depicted in Figure~\ref{fig:ablation}, clearly demonstrate the contribution of each component to enhancing performance of \ours. Particularly, the superiority of~\ours \\ ~over \ours-$\mathcal{L}_{\mathrm{alignment}}$, \ours-$\mathcal{L}_{\mathrm{prop}}$ and \ours-$\mathcal{L}_{\mathrm{mimic}}$ highlights indispensable roles of the alignment principle, the gradient matching strategy, and the preservation of graph properties when generating the new graph, respectively. The notable performance drop with \ours-$\mathcal{L}_{\mathrm{mimic}}$ underscores the critical importance of the loss $\mathcal{L}_{\mathrm{mimic}}$. In addition, the observed decrease in performance with \ours-init validates the effectiveness of our initialization strategy in improving graph transferability.

\begin{figure}[!t]
    \centering
    \includegraphics[width=1.0\columnwidth]{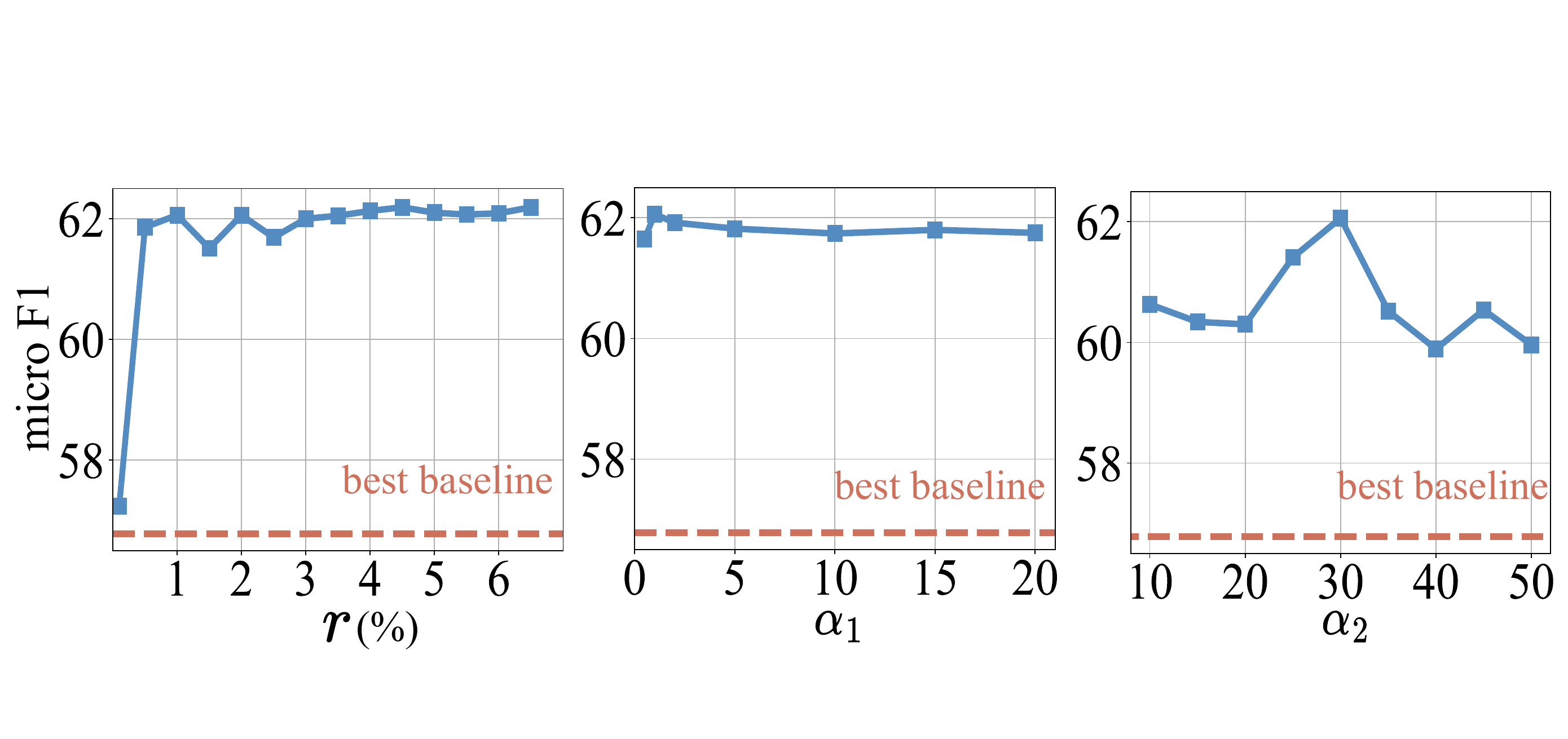}
    \vspace{-0.3in}
    \caption{
    Our results on D$\rightarrow$A task  w.r.t varying $r$, $\alpha_1$  and $\alpha_2$. The dashed line represents the performance of the best baseline.
    }~\label{fig:hyper-combine}
    \vspace{-0.1in}
\end{figure}

\vspace{-0.1in}
\vpara{Hyper-parameters analysis.}
Figure~\ref{fig:hyper-combine} shows the effects of the reduction rate $r$, the coefficients $\alpha_1$ for $\mathcal{L}_{\mathrm{alignment}}$  and $\alpha_2$ for $\mathcal{L}_{\mathrm{prop}}$, respectively. Regarding the reduction rate $r$, we observe that our model consistently outperforms the most competitive baseline model. This indicates our model's ability to scale down the graph while enhancing the transferability to the target graph. We also note that an excessively low reduction rate could hamper performance due to the difficulty in obtaining an informative yet overly small graph. Regarding $\alpha_1$ and $\alpha_2$, we find that both too small and large values could deteriorate performance, as a value too small would weaken the influence of $\mathcal{L}_{\mathrm{alignment}}$  and $\alpha_2$ for $\mathcal{L}_{\mathrm{prop}}$, while a value too large would negatively affect the balance with other loss components in \eqref{eq:opt-1}. Despite variations, our model consistently surpasses the best baseline, demonstrating that \ours~exhibits low sensitivity to hyper-parameter changes.

\vspace{-0.02in}
\vpara{Transferability under different levels of domain shifts.}
We further evaluate the transferability of our approach and baselines across different levels of domain shifts. We here focus on structural shifts and feature shifts between graphs. Specifically, we conduct experiments on a series of synthetic graphs constructed based on CSBM. {The source graph is constructed as $\operatorname{CSBM}(n=128, C^{\mathcal{S}}=[[0.6,0.3],[0.3,0.6]], \mu=1, \nu=-1)$ and remains fixed. For structural shift, the target graphs are defined by a sequence of CSBMs with parameters $n=128$, $\mu=1$, $\nu=-1$, and $C^{\mathcal T}$ randomly generated. The level of structural shift is quantified as $\| C^{\mathcal{T}}- C^{\mathcal{S}} \|_{F}$ and ranges from 0.05 to 0.45. As for feature shift, the target graphs are defined by a sequence of $\operatorname{CSBM}(n=128, C^{\mathcal{T}}=[[0.6,0.3],[0.3,0.6]], \mu=1+\Delta, \nu=-1+\Delta)$, where $\Delta$ varies in $[0.1, 0.9]$ and quantifies the level of feature shift.}

\RH{
The results of our model and the most competitive methods in the three categories (vanilla ERM, non-graph domain adaptation methods, and UGDA methods)  are shown in Table~\ref{tab:structural shift}. We observe a decrease in the performance of all methods as domain shift increases, while \ours~performs the best under various levels of domain shifts in most cases.  Besides, we find that \ours~is more effective under large domain shift, achieving more substantial gains. When there is a small domain divergence, the effectiveness of our method may not be as significant, and existing domain adaptation methods may achieve comparable performance.}

\begin{table}[!t]
\caption{Micro F1 scores under different levels of domain shift (structural shift and feature shift).}
\vspace{-0.1in}
\setlength{\tabcolsep}{1.5pt}
     \fontsize{9pt}{11pt}\selectfont
    \resizebox{1.0\columnwidth}{!}{
    \centering
     \renewcommand{\arraystretch}{1.1}
    \begin{tabular}{c|ccccc}
    \toprule[1.5pt]
        Structural shift & 0.05  & 0.15  & 0.25  & 0.35  & 0.45 \\ \hline
        ERM (GCN)  & 77.31(7.54)  & 68.72(5.83)  & 65.47(6.67)  & 65.50(6.97) & 56.38(5.10) \\ 
        CDAN & 77.31(7.66)  & 67.27(5.68)  & 67.56(8.00) & 68.68(7.08)  & 55.50(6.92) \\
        AdaGCN & 82.21(6.37)  & 77.62(7.18) & 72.58(7.63) & 70.73(7.87)  & 55.78(0.62) \\ \cdashline{1-6}[1pt/1pt]
        \ours & \textbf{83.34(4.26)} & \textbf{78.32(9.45)}  & \textbf{73.25(6.63)} & \textbf{72.54(8.86)} & \textbf{58.64(5.10)} \\ 
        Gain (\%) & +1.37 & +0.90  & +0.92 & +2.56 & +4.01\\ \midrule[1.5pt]
        Feature shift & 0.1 & 0.3 & 0.5 & 0.7 & 0.9 \\ \hline
        ERM (GCN) & 80.61(6.32) & 73.94(6.63) & 70.07(6.60) & 66.02(5.44) & 65.38(6.71) \\ 
        CDAN & 73.08(13.87) &75.35(7.05) &73.56(7.08) &63.83(11.03) &66.72(6.97) \\ 
        AdaGCN & \textbf{82.68(5.50)} & 77.73(6.40)& 76.67(8.05)& 66.33(7.35) & 67.52(8.66) \\ 
        \cdashline{1-6}[1pt/1pt]
        \ours & 82.47(6.22) &\textbf{78.21(6.28)} &  \textbf{77.34(3.75)} & \textbf{66.98(7.70)} &\textbf{68.46(5.90)} \\ 
        Gain (\%) & -0.25 & +0.61 & +0.87 & +0.98 & +1.39\\ \bottomrule[1.5pt]
    \end{tabular}}\label{tab:structural shift}
\end{table}

\vspace{-0.02in}
\vpara{Adopting \ours~with different GNNs with ERM and model-centric UGDAs.}
After obtaining the generated graph $G^\prime$,  it can be utilized to train diverse GNN architectures with ERM or fed into model-centric UGDA methods. The results are shown in Table \ref{table:erm-results}. We find that (1) using the generated graph $G^\prime$ to train different GNNs with ERM still exhibits superior performance compared to existing UGDA methods. This underscores the versatility and superior efficacy of our approach in bridging the gap between graphs from different domains; (2) by combining our method with established model-centric UGDA methods, further enhancements in performance are observed in some cases, notably in cross-domain scenarios like D $\rightarrow$ C and C $\rightarrow$ A.

\begin{table*}[h]
\caption{Micro F1 scores when adopting \ours~with different GNNs with ERM and model-centric UGDAs.  The \textbf{bold} numbers denote the best result. The asterisk ($\ast$) indicates that the result surpasses the most competitive baseline in Table 1.}
\vspace{-0.1in}
    \fontsize{9pt}{10pt}\selectfont
    \resizebox{1.9\columnwidth}{!}{
    \centering
    \renewcommand{\arraystretch}{1.3}
\begin{tabular}{c|cccccc} 
 \toprule[1pt]
Methods   & C$\rightarrow$D    & A$\rightarrow$D    & D$\rightarrow$C  & A$\rightarrow$C   & D$\rightarrow$A   & C$\rightarrow$A  \\ \hline
\ours+ERM~(GCN) &\textbf{72.56(0.61)}*&\textbf{69.65(0.26)}*&68.08(0.32)*&\textbf{75.61(0.24)}*&\textbf{62.06(0.68)}*&67.36(0.40)*\\ 
\ours+ERM~(GraphSAGE) & 70.73(0.41)*&68.97(1.28)*&67.70(0.48)*&	73.21(0.50)*&61.08(0.52)*	&66.79(0.22)*\\ 
\ours+ERM~(SGC) & 69.93(0.02)&66.23(0.21)&64.12(0.01)&70.56(0.28)	&58.65(0.01)*&65.42(0.03)*\\ 
\ours+ERM~(APPNP) & 68.22(0.05)&65.59(0.01)&61.89(1.70)&67.89(0.73)
&56.84(0.02)*&63.49(0.04)*\\ \hline
\ours+MMD & 71.13(0.32)*&	67.01(0.35)	
&68.05(0.42)*&	73.69(0.46)*&	60.57(0.61)*	&66.53(0.32)* \\ 
\ours+UDAGCN & 70.83(0.17)*&69.35(0.23)*&\textbf{68.31(0.32)}*&75.61(0.24)*&60.23(0.80)*&\textbf{67.40(0.47)}* \\ 
\bottomrule
\end{tabular}
}\label{table:erm-results}
\end{table*}

\vspace{-0.03in}
\vpara{Runtime comparison.}
Table~\ref{table:runtime} presents the runtime comparisons between \ours~and {the most competitive} baselines of three categories: vanilla ERM, Non-graph domain adaptation methods, and UGDA methods.  We observe that the runtime of \ours~ is comparable to vanilla ERM methods, yet it delivers superior performance.  This efficiency stems from the rescaled graph, which significantly reduces the time required for GNN training with ERM. We also note that the efficiency of our method facilitates the possibility of further applications, such as neural architecture search of GNNs and hyper-parameter optimization.

\begin{table}[t]
\caption{Runtime (sec) and Memory (Mb) comparison on Arxiv-degree. All the models are trained within 300 iterations and the reduction rate for \ours~is $0.25\%$.}
\vspace{-0.1in}
\centering
\setlength{\tabcolsep}{0.9pt}
     \renewcommand{\arraystretch}{1.0}
\begin{tabular}{c|ccc|c}
\toprule
&\multicolumn{3}{c}{\textbf{Runtime}} & \textbf{Memory} \\
 & generate $G^\prime$ & GNN training &total&  total \\ \hline
{ERM (GCN) }&- & 110.49  & 110.49&	2868\\
{CDAN} &- & 148.57& 148.57 &2868 \\
{AdaGCN} &- & 577.91  & 577.91&2956 \\
{\ours} &257.53& 18.57&  276.10& 2050 \\ 
\bottomrule
\end{tabular}~\label{table:runtime}
\end{table}
\vspace{-0.15in}
\section{Related Work}
\vspace{-0.1in}
\vpara{Unsupervised domain adaptation.} 
Unsupervised domain adaptation (UDA) aims to transfer knowledge from a labeled source domain to an unlabeled target domain and has demonstrated success in computer vision and natural language processing~\cite{venkateswara2017deep,ben2010theory}. Most existing work on UDA attempts to learn invariant representations across different domains~\cite{ajakan2014domain,zellinger2017central,long2015learning}. 

Recently, domain adaptation has been adapted to graph, and numerous studies have been conducted to address UGDA. These studies can be generally divided into two categories: one focuses on minimizing domain discrepancy metrics, while the other utilizes adversarial training techniques. The first class of methods aims to learn domain-invariant representations by minimizing pre-defined domain discrepancy metrics, such as class-conditional MMD~\cite{Shen_2021}, central moment discrepancy~\cite{Zellinger_2019}, spectral regularization~\cite{you2023graph}, graph subtree discrepancy~\cite{wu2023non}, tree mover’s distance~\cite{chuang2022tree}. In comparison, the adversarial training-based methods typically incorporate a domain classifier that adversarially predicts the domain of the representation. For instance, ~\citet{dai_graph_2022} and \citet{wu2020UDAGCN} utilize GNN models as feature extractors and train them in an adversarial manner to align the cross-domain distributions. \citet{qiao2023semi} further introduces shift parameters to enhance transferability to target graph during adversarial training. However, these methods consider UGDA from the model-centric perspective. Different from aforementioned works on UGDA, this paper addresses UGDA from a novel data-centric perspective that is allowed to modify the source graph.

To the best of our knowledge, the only UGDA method that modifies the source graph is a recent one ~\cite{liu2023structural}. But, their modification relies on the accurate estimation of target labels, and the process of estimating target labels is highly dependent on model-centric UGDA methods. In addition, their modification to the source graph is limited to edge weights.  In contrast, our method is purely data-centric and allows the modified graph to be directly utilized for training a GNN by ERM, achieving competitive performance without the need for sophisticated design. Besides, we enable modifications to graph structure, node features and node labels. We further investigate the impact of graph scale on the generalization bound and suggest that a smaller-scale modified graph can be effective.

\RH{Another line of research assume that source graph is not accessible during the adaptation process~\cite{mao2021source,jin2022empowering}. Specifically, \cite{mao2021source} focuses on enhancing the discriminative capability of the source model through structure consistency and information maximization. \cite{jin2022empowering}  explores modifying graph data during testing to improve model generalization and robustness. However, this setting is different from ours, and the absence of source graph poses difficulties in addressing shifts caused by differences in data distributions.}

\vpara{Data-centric AI.}
This recently introduced concept emphasizes the potential of data modification over model design~\cite{yang2023data,zha2023data,cao2023pre}. Subsequent works leverage the data-centric approach to address various questions in machine learning. Some works~\cite{xu2023better,jin2022empowering} take into account the impact of graph data on model generalization. \citet{xu2023better} considers the co-evolution of data and model to enhance data quality for pre-training without considering information in a downstream graph. \citet{jin2022empowering}  modifies test graph to address distribution shift at test time and then, the pretrained model is provided for inferences on the modified test graph. However, both of them overlook the rich information inherent in either the target graph or the source graph itself and, and thus fail to address UGDA.

Another line of research focuses on reducing graph scale. Several works have proposed methods for graph condensation or sparsiﬁcation while preserving information of the original graph~\cite{ cai2021graph,huang2021scaling}. However, these methods don't consider the transferability of the generated graphs, making them unsuitable for direct adoption in UGDA. \RH{In contrast, our method focuses on generating a new graph that can transfer effectively to the target domain, rather than merely retaining information.} Additionally, the process of reducing the graph scale is often sensitive to initialization, but there is currently no existing work that discusses this aspect.
\vspace{-0.25in}
\section{Conclusion}
This work investigates UGDA through a data-centric lens. Our analysis pinpoints the limitations in model-centric UGDA methods and shows the potential of data-centric methods for UGDA. We therefore introduce two data-centric principles for UGDA: the alignment principle and the rescaling principle, rooted in the generalization bound for UGDA. Guided by these principles, we propose \ours, a novel data-centric UGDA method. \ours~first generates a small yet transferable graph in replacement of the original training graph and trains a GNN on the newly generated graph with classic ERM setting. Numerical experiments demonstrate that \ours~achieves remarkable transferability to the target graph. 

\vspace{-0.1in}
\section*{Acknowledgements}
This work is supported by NSFC (62206056, 92270121), Zhejiang NSF (LR22F020005) and SMP-IDATA Open Youth Fund.

\bibliographystyle{ACM-Reference-Format}
\balance
\bibliography{reference}

\clearpage
\onecolumn
\appendix
\section{Appendix}
\subsection{Notations}\label{app:notation}
The main notations can be found in the following table.

\begin{table}[H]
\renewcommand{\arraystretch}{1.1}
\centering
 \begin{tabular}{lp{10cm}}
		\toprule 
		\textbf{Notation} & \textbf{Description}   \\
		\midrule
		$G^{\mathcal{S}}$, $G^{\mathcal{T}}$, $G^{\prime}$ & Source  graph, target  graph, and generated graph\\
            $ A^{\mathcal{S}}, X^{\mathcal{S}}, Y^{\mathcal{S}}$ 
            & Adjacency matrix, node feature matrix, and node label matrix of source  graph \\
            $ A^{\mathcal{T}}, X^{\mathcal{T}}, Y^{\mathcal{T}}$ 
            & Adjacency matrix, node feature matrix, and node label matrix of target  graph \\
         $ A^\prime, X^\prime, Y^\prime$ 
            & Adjacency matrix, node feature matrix, and node label matrix on generated graph \\
            $n^\mathcal{S}$, $n^\mathcal{T}$, $n^\prime$
            & Number of node in source graph, target graph and generated graph\\
            $\mathcal G, \mathcal Z, \mathcal Y$ &  Input space, representation space and label space \\
  		$f(\cdot)$, $g(\cdot)$  & Feature extractor and classifier \\
     	$\mathcal N$, $\boldsymbol{\mu}$, $\boldsymbol{\nu}$, $\boldsymbol{\tilde \mu}$, $\boldsymbol{\tilde \nu}$ & Normal distribution and its mean values \\
            $C$ & Probability matrix that is used to model edge connections \\
            $\epsilon, \hat \epsilon, \psi$ & Generalization error, empirical classification error, labeling function\\
            $\mathcal H$, $d$ & Hypothesis set and its pseudo-dimension  \\
            $\eta$ & Optimal combined error that can be achieved on both source and target graphs by the optimal hypothesis \\
            $g^{*}$, $f^{*}$ & Optimal hypothesis for joint error\\
            $W_1$ & Wasserstein distance \\
            $r$ & Reduction rate \\
            $\alpha_1$, $\alpha_2$ & Hyper-parameters for $\mathcal{L}_\mathrm{alignment}$, $\mathcal{L}_\mathrm{prop}$ \\
            $\rho$, $\phi$ & Function that transforms node features to graph structure, and its learnable parameter\\
            $\tau$  & Temperature hyper-parameter\\
            $L$, $D$ &  Normalized Laplacian and diagonal degree matrix \\
		\bottomrule 
	\end{tabular}
\caption{\label{tab:notations} Description of major notations.}
\vspace{-0.3in}
\end{table}

\subsection{Framework} \label{app:framework}

In this section, we detail the pseudocode for the algorithm behind \ours. We first describe the process for initializing the generated graph.

\begin{algorithm}[!h]
    \caption{Initialization of the generated graph}
    \begin{algorithmic}[1]
        \REQUIRE Source domain graph $G^{\mathcal{S}}=\big(A^{\mathcal{S}}, X^{\mathcal{S}}, Y^{\mathcal{S}} \big)$, target domain graph $G^{\mathcal{T}}=\big( A^\mathcal T, X^\mathcal T \big)$, reduction rate $r$, and the threshold hyper-parameter $\kappa$.
        \ENSURE Initialization for $X^{\prime}$ and  $Y^{\prime}$.
        \STATE Create $n^\prime$ plain nodes, where $n^\prime=\text{int}(rn^\mathcal S)$.
        \STATE Calculate $\tilde{A}=\operatorname{ReLU}(\operatorname{Sigmoid}\left(X^\mathcal S {X^{\mathcal S}}^{\top}\right)-\kappa)$.
        \STATE Calculate the number of nodes for each class in $G^\prime$ to ensure that the node labels $Y^\prime$ have the same distribution as $Y^\mathcal{S}$. Assign labels to plain nodes based on the above constraint.
        \STATE Initialize the candidates of node set $=\{ \}$.
        \FOR{range in $0,1,\cdots ,100$}
            \STATE Randomly select a set of $n^\prime$ nodes as $[n^\prime]$ and sample its ego-graph from graph based on $\tilde{A}$.
            \STATE Randomly select a set of nodes and sample its corresponding ego-graph from $G^\mathcal T$ .
            \STATE Compute the spectral distance $\Delta_\mathrm{spectral}$ based on the above two sets of ego-graphs.
            \STATE Add ($[n^\prime]$,  $\Delta_\mathrm{spectral}$) to the candidates of node set.
        \ENDFOR
        \STATE  Choose the $[n^\prime]$ that has the minimal $\Delta_\mathrm{spectral}$ within the  candidates of node set, and assign node features of $[n^\prime]$
        to the plain nodes as the initial values for $X^\prime$.
    \end{algorithmic}\label{alg:init}
\end{algorithm}
 
Then, we outline the overall procedure of  \ours.
\begin{algorithm}[!h]
    \caption{\ours}
    \begin{algorithmic}[1]
        \REQUIRE Source domain graph $G^{\mathcal{S}}=\big(A^{\mathcal{S}}, X^{\mathcal{S}}, Y^{\mathcal{S}} \big)$, target domain graph $G^{\mathcal{T}}=\big( A^\mathcal T, X^\mathcal T \big)$, and the number of iterations $T$.
        \ENSURE Trained GNN $h$.
        
        \STATE Initialize $X^\prime$ and $Y^{\prime}$ based on Algorithm~\ref{alg:init}.

        \STATE Pretrain the surrogate GNN model for $\mathcal{L}_\mathrm{alignment}$ by unsupervised loss~\cite{velivckovic2018deep}.
        \FOR{iteration in $1,2,\cdots ,T$}
            \STATE Train the surrogate GNN model for $\mathcal{L}_\mathrm{mimic}$ by supervised loss on $G^\prime$.
            \STATE Fix the parameters of surrogate models, and calculate the constraints $\mathcal{L}_\mathrm{mimic}$, $\mathcal{L}_\mathrm{alignment}$, and $\mathcal{L}_{\mathrm{prop}}$. 
            \STATE Update $X^\prime$ and $\phi$ by minimizing Eq. (\ref{eq:opt-1}).
        \ENDFOR
        \STATE Utilize $X^\prime$ and $\phi$ to generate $G^\prime$.
        \STATE Train GNN $h$ on $G^\prime$.
    \end{algorithmic}\label{alg:graphalign}
\end{algorithm}

\subsection{Addition Experimental Setup}\label{app:exp-setup}

\vpara{Description of baselines.} Below, we provide descriptions of the baselines used in the experiments.

\begin{itemize}[leftmargin=15pt]
\item \textbf{ERM (GIN)~\cite{xu2019powerful}, ERM (SAGE)~\cite{hamilton2017inductive}, ERM (GCN)~\cite{welling2016semi}}, 
train using GIN, GraphSAGE, and GCN under standard ERM.
\item \textbf{MMD}~\cite{long2015learning} and \textbf{CMD}~\cite{Zellinger_2019}, serve 
as classic baselines for non-graph domain adaptation. These works  regularize the model parameters by minimizing the distance of the representation mean and the correlation matrix of the representations between the source and target domains, respectively.
\item \textbf{DANN}~\cite{ajakan2014domain} and \textbf{CDAN}~\cite{long2018conditional} serve as classic baselines for non-graph domain adaptation. These works regularize the model parameters by blurring the classifier that is designed to discriminate between the source and target domains. 
\item \textbf{AdaGCN}~\cite{dai_graph_2022} is model-centric method for UGDA. It utilizes the techniques of adversarial domain adaptation with graph convolution, and employs the empirical Wasserstein distance between the source and target distributions of node representation as the regularization.
\item \textbf{UDAGCN}~\cite{wu2020UDAGCN} is a model-centric method for UGDA, which firstly combines adversarial learning with a GNN model.
\item \textbf{MIXUP}~\cite{han2022gmixup} is model-centric method for enhancing the out-of-distribution generalization ability, which augments graphs by interpolating features and labels between two random samples based on estimated graphon. 
\item \textbf{EERM}~\cite{wu2022eerm}  is model-centric method for enhancing the out-of-distribution generalization ability, which minimizes the variance of representations across different generated environments.
\item \textbf{SSReg} and \textbf{MFRReg}~\cite{you2023graph} is model-centric method which regularizes spectral properties to restrict the generalization risk bound.
\item \textbf{STRURW}~\cite{liu2023structural} modifies graph data by adjusting edge weights, while the modification relies on the training of data-centric UGDA methods.
\end{itemize}

\vpara{Detailed statistics for datasets.} Below is the statistics of the datasets used in our experiments.
\begin{table*}[!ht]
\fontsize{9pt}{10pt}\selectfont
\setlength \tabcolsep{2pt}
\renewcommand{\arraystretch}{1}
\centering
{
\begin{tabular}{lccccccccc}
 \toprule
  & DBLPv7    & Citationv1   & ACMv9 &$\text{DBLP}_\text{small}$&$\text{ACM}_\text{small}$& Cora-word& Cora-degree  & Arxiv-degre & Arxiv-time \\ \hline
\# Nodes  & 5,484 & 8,935 & 9,360 &5,578&7,410& 8213/3781 &9378/3454 & 102,474/66,868 &4,980/169,343 \\\
\# Edges &  8,130 &15,113 &15,602 &7,341&11,135&26,050/6,140 &54,452/488&   362,796/868,810& 6,103/1,166,243 \\
\# Features&6,775&6,775 &6,775&7,537&7,537& 8,710 &8,710  &128 &128\\
\# Classes   & 5&5 &5 &6 &6&70&70&40 &40\\           \bottomrule
\end{tabular}
}
\caption{
Dataset statistics. The slash (/) denotes the statistical information of the source and target datasets.}\label{tab:statistic}
\vspace{-0.2in}
\end{table*}

\vpara{Additional implementation details.} 
We further elaborate on the hyper-parameters used and the running environment. In the implementation of \ours, we set $\tau=0.07$, $\kappa=0.52$. As for the running environment, our model is implemented under the following software setting: Pytorch version 1.12.0+cu113, CUDA version 11.3, networkx version 2.3, torch-geometric version 2.3.0, sklearn version 1.0.2, numpy version 1.21.5, Python version 3.7.10. We conduct all experiments on the Linux system with an Intel Xeon Gold 5118 (128G memory) and a GeForce GTX Tesla P4 (8GB memory).

\vpara{Estimation of MMD distance.} We follow the standard procedure outlined by~\citet{pan2008transfer} to estimate $\mathrm{MMD}$:

\begin{equation}
\mathrm{MMD}^2(\mathbb P(f(G^{\prime})), \mathbb P(f(G^{\mathcal{T}}))) \approx \frac{1}{n^\prime(n^\prime-1)} \sum^{n^\prime}_i \sum^{n^\prime}_{j \neq i} k\left(x_i, x_j\right)+\frac{1}{n^\prime(n^\prime-1)} \sum^{n^\prime}_i \sum^{n^\prime}_{j \neq i} k\left(y_i, y_j\right) -\frac{2}{n^\prime n^\prime} \sum^{n^\prime}_i \sum^{n^\prime}_j k\left(x_i, y_j\right),
\end{equation}
where $k$ is the gaussian kernel defined by $k\left(\mathrm{x}, \mathrm{x}^{\prime}\right)=\mathrm{e}^{-\frac{\left\|\mathrm{x}-\mathrm{x}^{\prime}\right\|^2}{2 \sigma^2}}$, and $x$ and $y$ are node representation matrix on the $G^{\prime}$ and $G^{\mathcal{T}}$, respectively, provided by the surrogate model.  In practical, we sample an equivalent number of nodes $n^\prime$ from the target graph to compute the $\mathrm{MMD}$ to enhance the efficiency of estimation of the distribution of target graph.

\subsection{Additional Experimental Results}\label{app:exp}

\vpara{Performance and convergence under different initialization.} The initialization for $X^\prime$ is a crucial module that influences performance. We analyze the effectiveness from the perspectives of performance in UGDA and loss convergence.

To demonstrate the performance superiority of the initialization in tackling UGDA, we select the following baseline models for comparison: (1) Random: Random initialization. (2) K-Means~\cite{hartigan1979algorithm}: We first initialize $Y^\prime$ and determine the number of nodes for each class. For nodes belonging to a certain class, K-Means is applied to the  features of all nodes, dividing into $M$ clusters (where $M$ is the number of nodes for that class). Then, we randomly sample a node from each cluster and assign its features to plain nodes. (3) Herding~\cite{welling2009herding}:  Replace the above-mentioned K-Means with Herding for clustering. (4) K-Center~\cite{farahani2009facility}: Replace the above-mentioned K-Means with Herding for clustering.  (5) MMD: Replace the spectral distance $\Delta_\mathrm{spectral}$ in Algorithm~\ref{alg:init} with MMD distance. The results in Table~\ref{table:init-results} indicate the significance of our initialization in most cases.

\begin{figure}[h]     
    \centering
{\includegraphics[width=0.35\columnwidth]{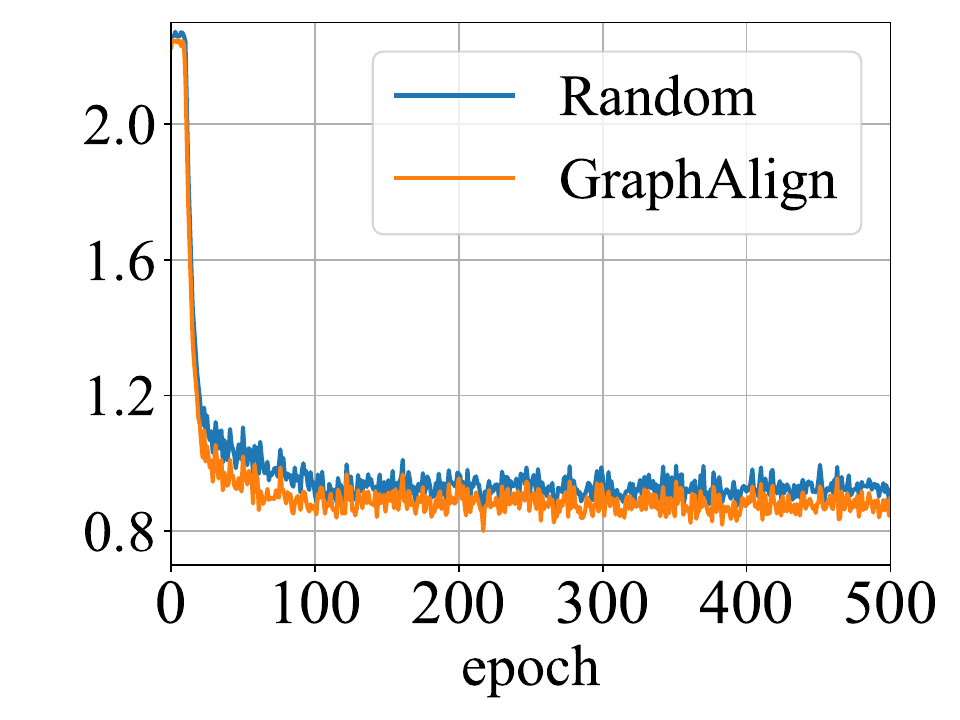}}
        \vspace{-0.15in}
    \caption{Loss curve between initialization of \ours~and random initialization on D$\rightarrow$A.
    }\label{fig:curve}
     \vspace{-0.2in}
\end{figure}

As for the impact of initialization on the loss convergence, Figure~\ref{fig:curve} illustrates a comparison between \ours~and a variant where \ours's specialized initialization is replaced with a random one. It is observed that utilizing the \ours~initialization significantly enhances the optimization process for $G^\prime$, resulting in improved convergence characterized by a reduced loss.

\begin{table*}[!h]
    \fontsize{9pt}{10pt}\selectfont
    \resizebox{0.5\columnwidth}{!}{
    \centering
    \renewcommand{\arraystretch}{1.3}
\begin{tabular}{c|cccccc} 
 \toprule[1pt]
Methods   & C$\rightarrow$D    & A$\rightarrow$D    & D$\rightarrow$C  & A$\rightarrow$C   & D$\rightarrow$A   & C$\rightarrow$A  \\ \hline
Random & 72.36 & 68.02 &67.03&72.61 &61.35& 67.01\\ 
K-Means & \textbf{72.59} & 68.51 &67.85 &73.95 &60.43& 66.85\\ 
Herding & 71.99 & \textbf{69.89} &67.79 &74.48 & 61.84 & 66.79\\ 
K-Center & 72.03 & 67.85 &67.44 &74.57 &61.42& 66.44\\ 
MMD & 71.99 & 67.61 &67.02 &74.46 &61.73& 66.90\\ \hline
\ours & 72.56 & 69.65 & \textbf{68.08} & \textbf{75.61} & \textbf{62.06} & \textbf{67.36} \\ 
\bottomrule
\end{tabular}
}
\caption{
  Micro F1 scores of different initialization methods. The \textbf{bold} numbers denote the best result.}\label{table:init-results}
  \vspace{-0.2in}
\end{table*}

\vpara{The impact of \ours~on the $\eta$.}  
{As discussed in \S\ref{subsec:bound}, $\eta$ in Eq. (\ref{thm:genbound}) impacts the generalization bound for UGDA. We here aim to investigate whether \ours~can help reduce $\eta$. The $\eta$ is calculated by the error ideal joint hypothesis $h$ when training on both source labeled graph $G^{\mathcal{S}}$ ($G^{\prime}$ for our method) and target labeled graph $G^{\mathcal{T}}$. The joint error is the sum of the error on both domains for the ideal joint hypothesis. }Table~\ref{tab:eta} compares the value of $\eta$ using the original source graph $G^{\mathcal{S}}$ and the generated graph $G^{\prime}$ obtained through our method. It can be observed that our approach effectively reduces the value of the $\eta$. This further highlights the superiority of \ours~in modifying data to address UGDA. 

\begin{table}[!h]
    \fontsize{9pt}{10pt}\selectfont
    \resizebox{0.65\columnwidth}{!}{ 
    \centering
    \renewcommand{\arraystretch}{1.3}
    \begin{tabular}{c|cccccc}
    \toprule[1pt]
       $\eta$ & A$\leftrightarrow$D & A$\leftrightarrow$C & D$\leftrightarrow$C & $\text{A}_\text{small} \leftrightarrow \text{D}_\text{small}$ & Cora-degree & Arxiv-degree \\ \hline
        $G^{\mathcal{S}}$ & 11.78 & 11.43 & 10.63 & 10.35 & 29.9 & 117.57\\ 
        $G^{\prime}$ & \textbf{3.87} & \textbf{3.39} & \textbf{3.35} & \textbf{4.58} & \textbf{18.3} & \textbf{51.57} \\ \bottomrule
    \end{tabular}}
    \caption{
  Comparisons of $\eta$ with different source graph ($G^{\mathcal{S}}$ or $G^{\prime}$). The \textbf{bold} numbers indicate the best result.}~\label{tab:eta}
  \vspace{-0.2in}
\end{table}

\vpara{Tranferability of \ours~across graphs with different homophily.} In this part, we evaluate the transferability under different domain shift from the homophily perspective. We further evaluate the transferability of our method and baselines across different graphs with different homophily. The homophily assumption is a common assumption in GNNs, and real-world graphs exhibit various of homophily. Exploring the performance of our method on datasets with different forms of homophily illustrate practical significance. We follow~\cite{barabasi1999emergence} to generate the target graph with different homophily ratio on Cora via preferential attachment. We fixed a graph with homophily set to 1.0 as the source graph, while the target graphs were generated with various homophily ratio ranging from 0.1 to 0.9. The results are shown in Table~\ref{exp:homo}.
We observe a decrease in performance as domain shift increased (lower homophily ratios). However, when encountering significant domain shift, we noticed that our model consistently achieved the best performance and exhibited relatively larger improvements. This indicates that our approach could effectively tackle more complex domain shift.

\begin{table}[!h]
 \fontsize{9pt}{10pt}\selectfont
    \resizebox{1.0\columnwidth}{!}{
    \centering
     \renewcommand{\arraystretch}{1.3}
    \begin{tabular}{c|cccccccccc}
    \toprule[1pt]
        Homophily ratio & 0.9 & 0.8 & 0.7 & 0.6 & 0.5 & 0.4 & 0.3 & 0.2 & 0.1 & 0 \\ \hline
        ERM & 97.80(0.06) & 93.91(0.16) & 88.42(0.08) & 85.02(0.52) & 77.55(0.46) & 68.91(0.19) & 65.76(0.81) & 60.18(0.54) & 58.01(0.28) & 57.20(0.81) \\
        MMD & 97.70(0.07) & 93.76(0.17) & 87.84(0.18) & 84.33(0.28) & 76.28(0.48) & 67.06(0.56) & 63.19(1.12) & 57.35(0.79) & 54.66(0.31) & 53.44(1.02) \\
        DANN & \textbf{97.82(0.00)} & \textbf{93.93(0.20)} &  88.34(0.09) & 85.10(0.45) & 77.55(0.49) & 68.88(0.34) & 65.69(0.86) & 60.49(0.47) & 58.28(0.20) & 57.23(0.52) \\ 
        UDAGCN & 96.02(0.88) & 89.85(2.64) & 88.87(3.12) & \textbf{85.28(4.93)} & 76.85(8.01) & 68.10(6.41) & 63.72(6.96) & 58.81(5.96) & 47.30(6.48) & 45.55(6.80) \\ \hline
        \ours & 97.72(0.07) & 93.62(0.16) & \textbf{88.93(0.61)} & 85.13(1.50) & \textbf{78.03(0.92)} & \textbf{69.93(0.43)} & \textbf{65.97(1.35)} & \textbf{60.86(0.72)} & \textbf{58.58(0.43)} & \textbf{57.65(0.53)} \\ \bottomrule
    \end{tabular}}
            \caption{
  Micro F1 scores under different level of homophily. The \textbf{bold} numbers denote the best result.}~\label{exp:homo}
  \vspace{-0.2in}
\end{table}

\begin{figure}[h]
    \centering
    \subfloat[{\footnotesize Target graph, UDAGCN} ]
    {\includegraphics[width=0.30\columnwidth]{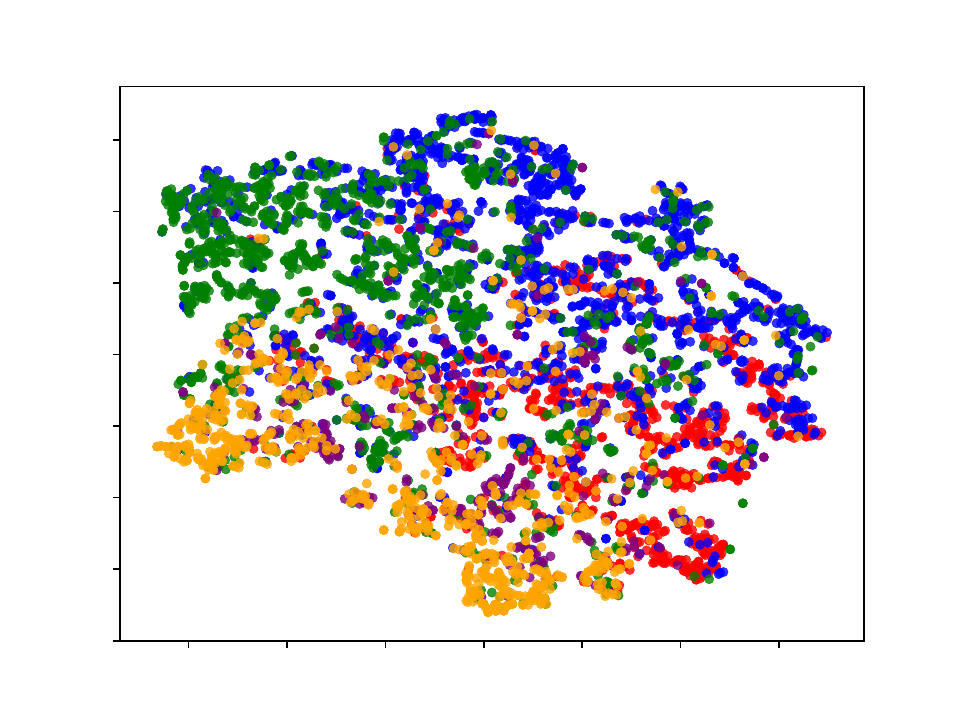}}
    \subfloat[{Target graph, \ours} ]
    {\includegraphics[width=0.30\columnwidth]{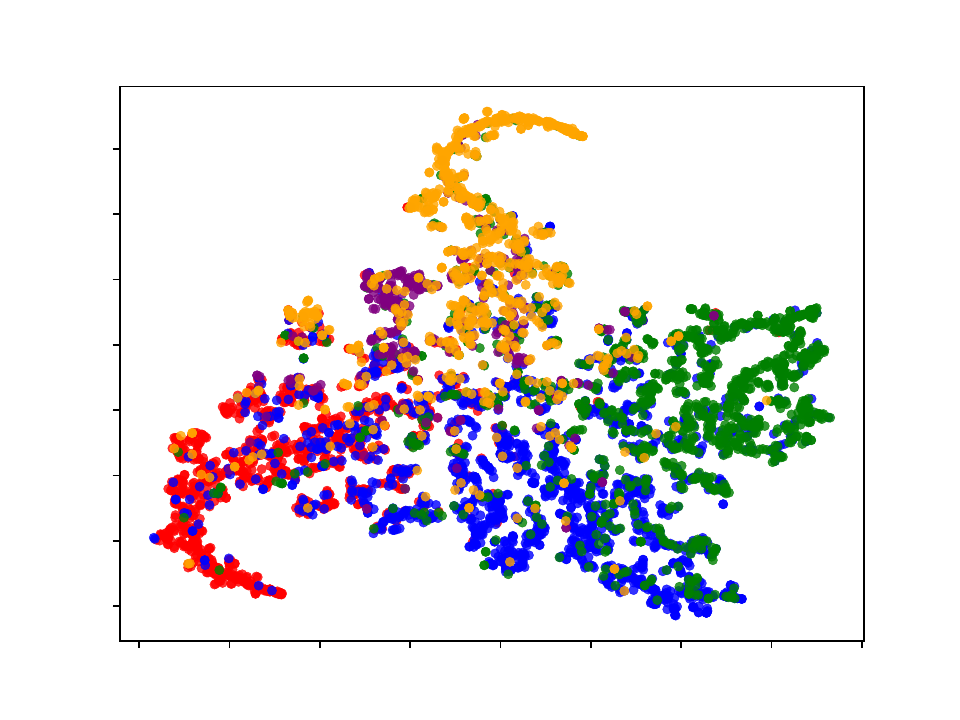}}
    \caption{  
    Visualization of node embeddings inferred by UDAGCN model and our method (\ours) for the A$\rightarrow$D task.
    }
    \label{fig:visual-emb}
\end{figure}

\vpara{Visualization for representation space.}
We delve into the comparison of the representation distributions generated by our model and UDAGCN, one of typical model-centric UGDA methods. We project the node representations of the target graph into a two-dimensional space using the t-SNE algorithm~\cite{van2008visualizing}. For this visualization, we set the perplexity parameter in t-SNE to 30 and assign different colors to nodes belonging to different classes, reflecting the representation distribution corresponding to distinct classes. 
It is apparent in Figure~\ref{fig:visual-emb} that the representations obtained through our method result in more concentrated clusters of nodes in the same class compared to UDAGCN. This indicates that training in the form of generating graphs can partially reduce domain-irrelevant noise and thus differentiating between distributions of different classes.

\subsection{Proofs}\label{app:proofs}
Here, we begin by presenting Example 1 and the proof of Proposition 1\&2, which addresses the inherent limitations in existing efforts on UGDA. Furthermore, we here provide the proof of generalization bound for UGDA in Theorem 1. Finally, we provide theoretical connections between
the property of the newly generated graph $G^\prime$ in Theorem 2.

\noindent {\textbf{Example 1.} 
Consider the source and target graphs generated by two CSBMs: $\operatorname{CSBM}(n, C^{\mathcal{S}},\boldsymbol{\mu},\boldsymbol{\nu})$ and $\operatorname{CSBM} (n, C^{\mathcal{T}}, \boldsymbol{\tilde{\mu}}, \boldsymbol{\tilde{\nu}})$, respectively. In both CSBMs, each class consists of $n/2$ nodes, and their edge connection probability
matrices are
\[
C^{\mathcal{S}} = \left[\begin{array}{cc}
    a & a \\
    a & a-\Delta
\end{array}\right], \qquad C^{\mathcal{T}} = \left[\begin{array}{cc}
    a-\Delta & a \\
    a & a
\end{array}\right],
\]
where $a$ and $\Delta$ are constants with $0 < \Delta < a < 1$.

\noindent {\textbf{Proof for Proposition 1.} We start the derivation as follows: Denote the  GNN model as $f$ and classifier $g$. Most of the previous domain adaptation methods primarily focused on learning domain-invariant representation to achieve domain adaptation, that is, the distribution of learnt embedding $\mathbb P(f(G^{\mathcal{S}}))=\mathbb P(f(G^{\mathcal{T}}))$.

Let's recall that the CSBM models for the source and target graph come with specific structures:
$$
C^{\mathcal{S}}=\left[\begin{array}{cc}
a & a \\
a & a-\Delta
\end{array}\right], \qquad C^{\mathcal{T}}=\left[\begin{array}{cc}
a-\Delta & a \\
a & a
\end{array}\right],
$$
In this context, $f$ is the most basic single-layer GNN, which typically employs message-passing mechanism for aggregation as follows:
$$
 h_v^{(k)}=\operatorname{COMBINE}\left(h_v^{k-1}, \operatorname{AGGREGATE}\left(\left\{h_{u}^{(k-1)} \mid u \in \mathcal{N}(v)\right\}\right) \right),
$$ 
where $k$ is the layer number, we omit the parameter with single-layer GNN model. From the above equation, we are aware that embeddings are impacted by the structural connections among neighbors as well as the corresponding features (without loss the generality, we let $\boldsymbol{\mu}=\boldsymbol{\tilde \mu}=\boldsymbol{\upsilon}=\boldsymbol{\tilde \upsilon}$ in CSBM). 
Therefore, the representation $h$ is dominated by $c_0$ and $c_1$, where $c_0$ or $c_1$ is the number of nodes in label 0 or 1. 
For simplicity, we denote the embedding space for label 0 as $\zeta_{0}(g,f)=\{ h \mid g \circ f (c_0,c_1)=0 ,h=f (c_0,c_1)\}$ and the same as embedding space $\zeta_{1}(g,f)=\{ h \mid g \circ f (c_0,c_1)=1 ,h=f (c_0,c_1)\}$ for label 1.

From the CSBM model, we can observe that the  edge probability among the node labelled with $\mathrm{Y}=0$ in source domain is equal to node labelled with  $\mathrm{Y}=1$ in the target domain, leading to the same neighbourhood structure. Therefore, no matter what model $f, g$ are chosen, $P\left[\zeta_{0}(g,f) \mid \mathrm{Y}^{\mathcal{S}}=0\right]=P\left[\zeta_{1}(g,f) \mid \mathrm{Y}^{\mathcal{T}}=1\right]$. Therefore, denote the classification error for graph encoder $f$ and classifier $g$ in domain ${\mathcal{D}}$ as $\epsilon_{\mathcal{D}}(g, f)$ as~\cite{shen2018wasserstein}. We have:

$$
\begin{aligned}
1=&P\left[\zeta_{0}(g,f) \mid \mathrm{Y}^{\mathcal{S}}=0\right]+ P\left[\zeta_{1}(g,f) \mid \mathrm{Y}^{\mathcal{S}}=0\right] \\
=&P\left[\zeta_{0}(g,f) \mid \mathrm{Y}^{\mathcal{T}}=1\right]+ P\left[\zeta_{1}(g,f) \mid \mathrm{Y}^{\mathcal{S}}=0\right] \\
\leq & 2\left(\epsilon^{\mathcal{T}}(g, f)+\epsilon^{\mathcal{S}}(g, f)\right) .
\end{aligned}
$$

The last inequality is because
$$
\begin{aligned}
\epsilon^{\mathcal{S}}(g, f)= & P\left[\zeta_{0}(g,f)  \mid  \mathrm{Y}^{\mathcal{S}}=1\right] P[ \mathrm{Y}^{\mathcal{S}}=1]+P\left[\zeta_{1}(g,f) \mid  \mathrm{Y}^{\mathcal{S}}=0\right] P[ \mathrm{Y}^{\mathcal{S}}=0] \\
& \geq \frac{1}{2} \max \left\{P\left[\zeta_{0}(g,f)  \mid  \mathrm{Y}^{\mathcal{S}}=1\right], P\left[\zeta_{1}(g,f) \mid  \mathrm{Y}^{\mathcal{S}}=0\right]\right\} .
\end{aligned}
$$
The above deduction is the same $\mathcal{T}$. In practical scenarios, models trained on seen data tend to perform well on seen data but poorly on unseen data. Therefore, based on practical considerations, we assume $\epsilon^{\mathcal{T}}(g, f) \geq \epsilon^{\mathcal{S}}(g, f)$. Therefore, under this assumption, we can prove $\epsilon^{\mathcal{T}}(g, f) \geq 0.25$.

\noindent {\textbf{Proposition 2.} Suppose that the feature extractor $f$ is a single-layer GNN, and shared between the source and target domains. Also suppose that a data-centric approach is employed to construct a new graph $G^\prime$ by modifying $G^\mathcal S$ with the constraint $\mathbb P(G^\prime) = \mathbb P(G^{\mathcal{T}})$. 
The GNN is trained with standard empirical risk minimization (ERM) on $G^\prime$, which minimizes the classification error on $G^\prime$. Then, there exists examples of graphs generated in Example~\ref{example} such that the classification error in the target domain is arbitrarily small.

\noindent {\textbf{Proof for Proposition 2.}  For simplicity but without losing generality, let's assume that the GNN encoder possesses the following characteristics: (1) Linearity. This property has already been demonstrated in some prior research~\cite{scarselli2008graph}; (2) During message propagation, the normalization step is applied. Based on these properties, we can represent the encoder as follows: $f\left(c_0,c_1 \right)=\left(c_0+c_1\right) / n$. 

Let's explore the constraint $\mathbb P(G^{\prime})=\mathbb P(G^{\mathcal{T}})$ by the divergence of node representations. Intuitively, when $\mathbb P(G^{\prime})=\mathbb P(G^{\mathcal{T}})$ is satisfied, it implies that for nodes with the same label, their nodes representation distributions have the same expectation.  The expectation of the representation distribution can be expressed as follows (let's consider nodes of category 0):
$$
\mathbb{E}(h^{\prime})= a, \mathbb{E}(h^{\mathcal{T}})=a-\frac{\Delta}{2}.
$$
Therefore, the divergence of node representations can be calculated by $L_\text{distance}=\operatorname{dis}(\mathbb{E}(h^{\prime}), \mathbb{E}(h^{\mathcal{T}}) )$, where $\operatorname{dis}(\cdot,\cdot)$ is the distance metrics. Here, we choose the most common Frobenius norm function as our distance function, 
$
L_\text{distance}=\operatorname{dis}( \mathbb{E}(h^{\prime}) - \mathbb{E}(h^{\mathcal{T}}) ) =\|a-(a-\frac{\Delta}{2}) \|_{F} =\frac{\Delta}{2}.
$ It's important to recall that we have two distinct cases as follows (where $\operatorname{B}(\cdot)$ and $\operatorname{Bern}(\cdot)$ represent the binomial distribution and Bernoulli distribution):

\begin{itemize}
\item If $v$ is from class 0 in the source domain, $c_0 \sim \operatorname{B}(n / 2,a) , c_1 \sim \operatorname{B}(n / 2, a)$.
\item If $v$ is from class 0 in the target domain, $c_0 \sim \operatorname{B}(n / 2,a-\Delta) , c_1 \sim \operatorname{B}(n / 2, a)$.
\end{itemize}
As $c_1$ and $c_0$ are always independent, if $v$ is from class 0 in the target domain, the node e $h=\frac{1}{n}\left(\sum_{i=1}^{n / 2} Z_i-\sum_{i=1}^{n / 2} Z_i^{\prime}\right)$, where $Z_i \sim \operatorname{Bern}(a)$ and $Z_i^{\prime} \sim \operatorname{Bern}(a)$, and all $Z_i$ 's and $Z_i^{\prime}$ 's are independent. Therefore, using Hoeffding's inequality, we have
$$
P\left(h-\mathbb{E}\left[h\right]>t\right) \leq \exp \left(-\frac{n t^2}{2}\right).
$$

Defining the classifier as $g(h) = 0$ when $h < x$ or $g(h) = 1$ when $h > x$. Here, $x$ is a constant. Under this classifier, we can find that the classification error for node with label 0 in source domain is $\exp \left(-\frac{n x^2}{2}\right)$. By setting $t=x$ and $\mathbb{E}\left[h\right]=0$ for node with label 0 in source domain, $P\left(h> x\right) \leq \exp \left(-\frac{n x^2}{2}\right)$. Similarly, the classification error for node with label 1 in source domain is $\exp \left(-\frac{n (x-\frac{\Delta}{2})^2}{2}\right)$. In the context of ERM, the error on the source is minimized. Therefore, we optimize the optimal classifier based on $\frac{1}{2}(\exp \left(-\frac{n x^2}{2}\right)+\exp \left(-\frac{n (x-\frac{\Delta}{2})^2}{2}\right))$. By analyzing the extreme points of the function, we can determine that the optimal classifier is given by setting $x=\frac{\Delta}{4}$.
After using such a classifier in the target domain, we can similarly achieve an classification error of $\epsilon^{\mathcal{T}} \leq 1-\exp \left(-\frac{n \Delta^2}{32}\right)$. 
It shows that as $\Delta$ decreases,  the classification error $\epsilon^{\mathcal{T}}$ becomes smaller. And, the distance function loss can serve as an upper bound to optimize and reduce the error
$$
\epsilon^{\mathcal{T}} \leq 1-\exp \left(-\frac{n \Delta^2}{32}\right)  = 1-\exp \left(-\frac{n L_\text{distance}^2}{8}\right).
$$

Therefore, there exist cases that classification error in the target domain can approach 0 by adopting a data-centric method.

\noindent {\textbf{Theorem 1.}  Denote by $L_\mathrm{GNN}$ the Lipschitz constant of the GNN model $g \circ f$. Let the hypothesis set be $\mathcal{H} = \left\{h = g \circ f : \mathcal G \rightarrow \mathcal Y \right\}$, and let the pseudo-dimension be $\operatorname{Pdim}(\mathcal{H}) = d$. The following inequality holds with a probability of at least $1-\delta$:

\begin{equation}
\begin{aligned}
\epsilon^{\mathcal{T}}(g, f) &\leq \hat{\epsilon}^{\mathcal{S}}(g, f) + \eta
+ \underbrace{2 L_{\mathrm{GNN}} W_1\big(
{\mathbb P(G^{\mathcal{S}})}, \mathbb P(G^{\mathcal{T}})\big)}_{\text{alignment term}} \\
&\quad \mathrel{+} \underbrace{\sqrt{\frac{4 d}{{n^{\mathcal{S}}}} \log \left(\frac{e {n^{\mathcal{S}}}}{d}\right)+\frac{1}{
{n^{\mathcal{S}}}} \log \left(\frac{1}{\delta}\right)}}_{\text{rescaling term}}, 
\end{aligned}~\label{eq:bound-proof}
\end{equation}
where $\hat \epsilon^{\mathcal S} (g,f) = ({1}/{n^\mathcal S}) \| g \circ f(G^{\mathcal{S}}) -\psi^{\mathcal{S}}(G^{\mathcal{S}}) \|$ is the empirical classification error in source domain with $\psi^\mathcal S$ the true labeling function on the source domain, $\eta = \min_{h \in \mathcal H} \big\{\epsilon^{\mathcal{S}}(g^{*}, f^{*}) + \epsilon^{\mathcal{T}}(g^{*}, f^{*})\big\}$ denotes the optimal combined error that can be achieved on both source and target graphs by the optimal hypothesis $g^{*}$ and $f^{*}$, $\mathbb P(G^{\mathcal{S}})$ and $\mathbb P(G^{\mathcal{T}})$ are the graph distribution of source and target domain respectively, the probability distribution $\mathbb P(G)$ of a graph $G$ is defined as the distribution of all the ego-graphs of $G$, and $W_1(\cdot, \cdot)$ is the Wasserstein distance.

\noindent {\textbf{Proof for Theorem 1.}  We first introduce the following inequality to be used that:
$$
\begin{aligned} \epsilon^{\mathcal{T}}(g, f)  & \leq \epsilon^{\mathcal{T}}(g^{*}, f^{*})+\epsilon^{\mathcal{T}}(g,f|g^{*}, f^{*}) \\ & =\epsilon^{\mathcal{T}}(g^{*}, f^{*})+\epsilon^{\mathcal{S}}(g,f|g^{*}, f^{*})+\epsilon^{\mathcal{T}}(g,f|g^{*}, f^{*})-\epsilon^{\mathcal{S}}(g,f|g^{*}, f^{*}).
\end{aligned}
$$

Here, we assume that the Lipschitz constant for the GNN model $g \circ f$ is denoted as $L_{\mathrm{GNN}}$. We denote $\epsilon^{\mathcal{T}}(g,f|g^{*}, f^{*})$  as   $\small \epsilon^{\mathcal{T}}(g,f|g^{*}, f^{*})= \mathbb{E}_{\mathbb P(G^{\mathcal{T}})} \left( \| g \circ f(G^{\mathcal{T}}) -g^{*} \circ f^{*}(G^{\mathcal{T}}) \| \right)$ and $\small \epsilon^{\mathcal{S}}(g,f|g^{*}, f^{*})= \mathbb{E}_{\mathbb P(G^{\mathcal{S}})}  \left( \| g \circ f(G^{\mathcal{S}}) -g^{*}\circ f^{*}(G^{\mathcal{S}}) \| \right)$.

According to Lemma 1 from~\cite{shen2018wasserstein}, we proof the following equation:
$$
\begin{aligned} \epsilon^{\mathcal{T}}(g, f)  & \leq \epsilon^{\mathcal{T}}(g^{*}, f^{*})+\epsilon^{\mathcal{S}}(g,f|g^{*}, f^{*})+\epsilon^{\mathcal{T}}(g,f|g^{*}, f^{*})-\epsilon^{\mathcal{S}}(g,f|g^{*}, f^{*})\\
& \leq \epsilon^{\mathcal{T}}(g^{*}, f^{*})+\epsilon^{\mathcal{S}}(g,f|g^{*}, f^{*})+ 2 L_{\mathrm{GNN}}  W_1\left(\mathbb P(G^{\mathcal{S}}),\mathbb P(G^{\mathcal{T}})\right)\\ & \leq \epsilon^{\mathcal{T}}(g^{*}, f^{*})+\epsilon^{\mathcal{S}}(g, f)+\epsilon^{\mathcal{S}}(g^{*}, f^{*})+ 2 L_{\mathrm{GNN}}  W_1\left(\mathbb P(G^{\mathcal{S}}),\mathbb P(G^{\mathcal{T}})\right)\\ & =\epsilon^{\mathcal{S}}(g, f)+2 L_{\mathrm{GNN}}  W_1\left(\mathbb P(G^{\mathcal{S}}),\mathbb P(G^{\mathcal{T}})\right)+\eta.
\end{aligned}
$$

We next link the bound with the empirical risk and labeled sample size by showing, with probability at least $1-\delta$ that:
$$
\begin{aligned}
\epsilon^{\mathcal{T}}(g, f) \leq & \epsilon^{\mathcal{S}}(g, f)+2 L_{\mathrm{GNN}}  W_1\left(\mathbb P(G^{\mathcal{S}}),\mathbb P(G^{\mathcal{T}})\right)+\eta \\
\leq & \hat{\epsilon}^{\mathcal{S}}(g, f)+2 L_{\mathrm{GNN}}  W_1\left(\mathbb P(G^{\mathcal{S}}),\mathbb P(G^{\mathcal{T}})\right)  +\sqrt{\frac{2 d}{n^{\mathcal{S}}} \log \left(\frac{e n^{\mathcal{S}}}{d}\right.})+\sqrt{\frac{1}{2 n^{\mathcal{S}}} \log \left(\frac{1}{\delta}\right)} +\eta.
\end{aligned}
$$

With the assistance of the Cauchy-Schwarz inequality, we can further derive the following inequality and provide the final conclusion.
$$
\begin{aligned}
\epsilon^{\mathcal{T}}(g, f) \leq & \hat{\epsilon}^{\mathcal{S}}(g, f)+2 L_{\mathrm{GNN}}  W_1\left(\mathbb P(G^{\mathcal{S}}),\mathbb P(G^{\mathcal{T}})\right)  +\sqrt{\frac{2 d}{n^{\mathcal{S}}} \log \left(\frac{e n^{\mathcal{S}}}{d}\right.})+\sqrt{\frac{1}{2 n^{\mathcal{S}}} \log \left(\frac{1}{\delta}\right)} +\eta \\
\leq & \hat{\epsilon}^{\mathcal{S}}(g, f)+2 L_{\mathrm{GNN}}  W_1\left(\mathbb P(G^{\mathcal{S}}),\mathbb P(G^{\mathcal{T}})\right)  +\sqrt{2} \sqrt{\frac{2 d}{n^{\mathcal{S}}} \log \left(\frac{e n^{\mathcal{S}}}{d}\right)+\frac{1}{2 n^{\mathcal{S}}} \log \left(\frac{1}{\delta}\right)} +\eta\\
= & \hat{\epsilon}^{\mathcal{S}}(g, f)+2 L_{\mathrm{GNN}}  W_1\left(\mathbb P(G^{\mathcal{S}}),\mathbb P(G^{\mathcal{T}})\right)  + \sqrt{\frac{4 d}{n^{\mathcal{S}}} \log \left(\frac{e n^{\mathcal{S}}}{d}\right)+\frac{1}{n^{\mathcal{S}}} \log \left(\frac{1}{\delta}\right)} +\eta.
\end{aligned}
$$

\noindent {\textbf{Theorem 2.} Let $G^\prime$ and  $G^{\mathcal{T}}$ be the newly generated graph and the target graph. Given a GNN graph encoder~$f$, the transferability of the GNN $f$ satisfies
\begin{equation}
\left\|f(G^{\prime})-f(G^{\mathcal{T}})\right\|_2
\leq \xi_1
\Delta_\mathrm{spectral}\left(G^{\prime}, G^{\mathcal{T}}\right)+\xi_2,
\end{equation}
where $\xi_1$ and $\xi_2$ are two positive constants, and $\Delta_\mathrm{spectral}\left(G^{\prime}, G^{\mathcal{T}}\right) = \tfrac{1}{n^{\prime}n^{\mathcal{T}}} \sum_{i=1}^{n^{\prime}} \sum_{j=1}^{n^{\mathcal{T}}} \|L_{G^\prime_i}-L_{G^{\mathcal T}_j} \|_2$ measures the spectral distance between $G^{\prime}$ and $G^{\mathcal{T}}$. Here $G^\prime_i$ is the ego-graph of node $i$ in $G^{\prime}$, and $L_{G^\prime_i}$ is its normalized graph Laplacian. The graph Laplacian $L_{G^{\mathcal T}_j}$ is defined in a similar manner.

\noindent \textbf{Proof of Theorem 2.}
Consider a graph encoder $f$ (usually instantiated as a GNN) with $k$ layers and $1-$hop graph filter $ \Lambda(L)$. Our focus lies on the central node's representation, which is acquired through a $k$-layer GCN utilizing a 1-hop polynomial filter $\Lambda (L)=I d-L$. This particular GNN model is widely employed in various applications. We denote the representations of nodes $i$ for all $i=1, \cdots, n$ in the final layer of the GCN, taking a node-wise perspective:
$
Z_i^{(k)}=\sigma\left(\sum_{j \in \mathcal{N}\left(i\right)} a_{i j} Z_j^{(k-1)^T} \theta^{(k)}\right) \in \mathbb{R}^d,$
where $a_{i j}=[\Lambda (L)]_{i j} \in \mathbb{R}$ is the weighted link between node $i$ and $j$; and $\theta^{(k)} \in \mathbb{R}^{d \times d}$ is the weight for the $k$-th layer sharing across nodes. Then $\theta=\left\{\theta^{(\ell)}\right\}_{\ell=1}^k$. We may denote $Z_i^{(\ell)} \in \mathbb{R}^d$ similarly for $\ell=1, \cdots, k-1$, and $Z_i^0=X_i \in \mathbb{R}^d$ as the node feature of center node $i$. With the assumption of GCN in the statement, we consider that only the $k-$hop ego-graph $G^\prime_i$ centered at $X_i$ is needed to compute $Z_i^{(k)}$ for any $i=1, \cdots, n$.

Denote $L_{G^\prime_i}$ as the out-degree normalised graph Laplacian of $G^\prime_i$, which is defined with respect to the direction from leaves to the centre node in $G^\prime_i$. We write the $\ell$-th layer representation as follows
$$
\left[Z_i^{(\ell)}\right]_{k-\ell+1}=\sigma\left(\left[\Lambda\left(L_{G^\prime_i}\right)\right]_{k-\ell+1}\left[Z_i^{(\ell-1)}\right]_{k-\ell+1} \theta^{(\ell)}\right).
$$

Assume that $\forall i$, $ \max _{\ell}\left\|Z_i^{(\ell)}\right\|_2 \leq c_z $, and $\max _{\ell}\left\|\theta^{(\ell)}\right\|_2 \leq c_\theta$. Suppose that the activation function $\sigma$ is $\rho_\sigma$-Lipschitz function. Then, for $\ell=1, \cdots, k-1$, we have
$$ 
\begin{aligned}
&\left\|\left[Z_{i}^{(\ell)}\right]_{k-\ell}-\left[Z_{i^{\prime}}^{(\ell)}\right]_{k-\ell}\right\|_2 \\
&\leq\|[\sigma\left(\left[\Lambda\left(L_{G^\prime_i}\right)\right]_{k-\ell+1}\left[Z_i^{(\ell-1)}\right]_{k-\ell+1} \theta^{(\ell)}\right) \\
&-\sigma\left(\left[\Lambda\left(L_{G^{\mathcal T}_j}\right)\right]_{k-\ell+1}\left[Z_{i^{\prime}}^{(\ell-1)}\right]_{k-\ell+1} \theta^{(\ell)}\right)]_{k-\ell}) \|_2 \\
& \leq \rho_\sigma c_\theta\left\|\Lambda\left(L_{G^\prime_i}\right)\right\|_2\left\|\left[Z_i^{(\ell-1)}\right]_{k-\ell+1}-\left[Z_{i^{\prime}}^{(\ell-1)}\right]_{k-\ell+1}\right\|_2 \\
&+\rho_\sigma c_\theta c_z\left\|\Lambda\left(L_{G^\prime_i}\right)-\Lambda\left(L_{G^{\mathcal T}_j}\right)\right\|_2.
\end{aligned}
$$
Since $\left[\Lambda\left(L_{G^\prime_i}\right)\right]_{k-\ell+1}$ is the principle submatrix of $\Lambda\left(L_{G^\prime_i}\right)$. We equivalently write the above equation as $A_{\ell} \leq a A_{\ell-1}+b$, where $a$ and $b$ are the coefficient. And we have
$$
\begin{aligned}
A_{\ell} &\leq  a A_{\ell-1}+b \leq  a^{2} A_{\ell-2}+b(1+a) \leq \dots \\
&\leq a^{\ell} A_0+\frac{a^{\ell}-1}{a-1} b.\\
\end{aligned}
$$

Therefore, for any $\ell=1, \cdots, k$, we have an upper bound for the hidden representation difference between $G^\prime_i$ and $G^{\mathcal T}_j$ by substitute coefficient $a$ and $b$,
$$
\begin{aligned}
\left\|\left[Z_i^{(\ell)}\right]_{k-\ell}-\left[Z_{i^{\prime}}^{(\ell)}\right]_{k-\ell}\right\|_2
& \leq\left(\rho_\sigma c_\theta\right)^{\ell}\left\|\Lambda\left(L_{G^\prime_i}\right)\right\|_2^{\ell}\left\|\left[X_i\right]-\left[X_{i^{\prime}}\right]\right\|_2 \\
& +\frac{\left(\rho_\sigma c_\theta\right)^{\ell}\left\|\Lambda\left(L_{G^\prime_i}\right)\right\|_2^{\ell}-1}{\rho_\sigma c_\theta\left\|\Lambda\left(L_{G^\prime_i}\right)\right\|_2-1} \rho_\sigma c_\theta c_z\left\|\Lambda\left(L_{G^\prime_i}\right)-\Lambda\left(L_{G^{\mathcal T}_j}\right)\right\|_2 .
\end{aligned}
$$
Specifically, for $\ell=k$, we obtain the upper bound for center node representation $ \left\|\left[Z_i^{(k)}\right]_0-\left[Z_{i^{\prime}}^{(k)}\right]_0\right\| \equiv$ $\left\|Z_i-Z_{i^{\prime}}\right\|$. Assuming that the difference in features between any two nodes does not exceed a constant, namely,
$\left\|\left[X_i\right]-\left[X_{i^{\prime}}\right]\right\|_2 \leq c_x$.
Suppose that $\forall i$, $\left\|\Lambda\left(L_{G^\prime_i}\right)\right\|_2 \leq c_L$ 
because that graph Laplacians are normalised. Since $\Lambda$ is a linear function for $L$, We have 
$$
\begin{aligned}
\left\|Z_i-Z_{i^{\prime}}\right\|_2 & \leq \left(\rho_\sigma c_\theta c_L\right)^k c_x + \frac{\left(\rho_\sigma c_\theta c_L\right)^{k}-1}{\rho_\sigma c_\theta c_L-1} 
c_\theta c_z\left\|\Lambda\left(L_{G^\prime_i}\right)-\Lambda\left(L_{G^{\mathcal T}_j}\right)\right\|_2 \\
& \leq  \chi_1 \left\|L_{G^\prime_i}-L_{G^{\mathcal T}_j}\right\|_2+\chi_2, 
\end{aligned}
$$
where $\chi_1=\frac{\left(\rho_\sigma c_\theta c_L\right)^{k}-1}{\rho_\sigma c_\theta c_L-1} c_\theta c_z$ and $\chi_2= \left(\rho_\sigma c_\theta c_L\right)^k c_x $.

Therefore, let's rephrase the following equation.
$$
\begin{aligned}
\left\|f(G^{\prime})-f(G^{\mathcal{T}})\right\|_2 \leq \frac{\chi_1}{n^{\prime}n^{\mathcal{T}}}  \sum_{i=1}^{n^{\prime}} \sum_{j^{\prime}=1}^{n^{\mathcal{T}}} \left\|L_{G^\prime_i}-L_{G^{\mathcal T}_j}\right\|_2+\chi_2. 
\end{aligned}
$$
Finally, let $\xi_1= \chi_1$ and $\xi_2= \chi_2$, concluding the proof.

\subsection{Theoretical connections between Wasserstein distance and MMD distance}\label{app:add-proofs}

Building upon previous work~\cite{zhang2021convergence,redko2020survey,feydy2019interpolating}, in this section, we establish a connection between Wasserstein distance and MMD distance. This demonstrates that MMD distance can provide a bounded interval for the Wasserstein distance.

\begin{theorem}
The Maximum Mean Discrepancy (MMD) distance provides a bounded interval for the Wasserstein distance. Given two graph $G_1$ and $G_2$ we have
$$
{\mathrm{MMD}}\left(\mathbb P(G_1), \mathbb P(G_2)\right) \leq  W_1\left(\mathbb P(G_1), \mathbb P(G_2)\right) \leq \sqrt{{\mathrm{MMD}}^2\left(\mathbb P(G_1), \mathbb P(G_2)\right)+\tau},
$$
where $\tau = \left\|\mu\left[\mathbb P(G_1)\right]\right\|_{\mathcal{H}} + \left\|\mu\left[\mathbb P(G_2)\right]\right\|_{\mathcal{H}}$, $\mathcal{H}$ is a RKHS space and $\mu$ is the mean value.
\end{theorem}

\noindent {\textbf{Proof for Theorem 3.}
To quantify various distances, we first define the MMD distance and Wasserstein distance. Given two probability measures $\mathbb P(G_1)$ and $\mathbb P(G_2)$ defined on a measurable space $\mathbf{X}$, let $\mathcal{F}=\left\{f:\|f\|_{\mathit{H}} \leq 1\right\}$ where $\mathit{H}$ is a RKHS space. Then, the MMD distance is defined as follows:
$$
\mathrm{MMD}\left(\mathbb P(G_1), \mathbb P(G_2)\right)=\sup _{\|f\|_{\mathit{H}} \leq 1}\left|\int f d\left(\mathbb P(G_1)-\mathbb P(G_2)\right)\right|.
$$

And Wasserstein distance can be defined as follows:
$$
W_1(\mathbb P(G_1), \mathbb P(G_2))=\inf _{\gamma \in \Gamma(\mathbb P(G_1), \mathbb P(G_2))} \int_{\mathbf{X} \times \mathbf{X}} c\left(\mathbf{x}, \mathbf{x}^{\prime}\right)  d \gamma\left(\mathbf{x}, \mathbf{x}^{\prime}\right) ,
$$

To derive this link between the Wasserstein distances and MMD distances, we first assume that the cost function in the Wasserstein distance is $c\left(\mathbf{x}, \mathbf{x}^{\prime}\right)=\left\|\phi(\mathbf{x})-\phi\left(\mathbf{x}^{\prime}\right)\right\|_{\mathit{H}}$ ($\phi: \mathbf{X} \rightarrow \mathit{H}$), then the following results can be obtained:
$$
\begin{aligned}
\left\|\int_{\mathbf{X}} f d\left(\mathbb P(G_1)-\mathbb P(G_2)\right)\right\|_{\mathit{H}} & =\left\|\int_{\mathbf{X} \times \mathbf{X}}\left(f(\mathbf{x})-f\left(\mathbf{x}^{\prime}\right)\right) d \gamma\left(\mathbf{x}, \mathbf{x}^{\prime}\right)\right\|_{\mathit{H}} \\
& \leq \int_{\mathbf{X} \times \mathbf{X}}\left\|f(\mathbf{x})-f\left(\mathbf{x}^{\prime}\right)\right\|_{\mathit{H}} d \gamma\left(\mathbf{x}, \mathbf{x}^{\prime}\right) \\
& =\int_{\mathbf{X} \times \mathbf{X}}\left\|\langle f(\mathbf{x}), \phi(\mathbf{x})\rangle-\left\langle f\left(\mathbf{x}^{\prime}\right), \phi\left(\mathbf{x}^{\prime}\right)\right\rangle\right\|_{\mathit{H}} d \gamma\left(\mathbf{x}, \mathbf{x}^{\prime}\right) \\
& \leq\|f\|_{\mathit{H}} \int_{\mathbf{X} \times \mathbf{X}}\left\|\phi(\mathbf{x})-\phi\left(\mathbf{x}^{\prime}\right)\right\|_{\mathit{H}} d \gamma\left(\mathbf{x}, \mathbf{x}^{\prime}\right).
\end{aligned}
$$
Now taking the supremum over $f$ and the infimum over $\gamma \in \Pi\left(\mathbb P(G_1), \mathbb P(G_2)\right)$, this gives
$
\mathrm{MMD}\left(\mathbb P(G_1), \mathbb P(G_2)\right) \leq W_1\left(\mathbb P(G_1), \mathbb P(G_2)\right).
$

This result holds under the hypothesis that $c\left(\mathbf{x}, \mathbf{x}^{\prime}\right)=\left\|\phi(\mathbf{x})-\phi\left(\mathbf{x}^{\prime}\right)\right\|_{\mathit{H}}$. On the other hand, in~\cite{gao2014minimum}, the authors showed that $W_1\left(\mathbb P(G_1), \mathbb P(G_2)\right)$ with this particular ground metric can be further bounded, as follows $
W_1\left(\mathbb P(G_1), \mathbb P(G_2)\right) \leq \sqrt{\mathrm{MMD}^2\left(\mathbb P(G_1), \mathbb P(G_2)\right)+\tau},
$ where $\tau=\left\|\mu\left[\mathbb P(G_1)\right]\right\|_{\mathit{H}}+\left\|\mu\left[\mathbb P(G_2)\right]\right\|_{\mathit{H}}$. Therefore, the overall inequality
\[
\mathrm{MMD}\left(\mathbb P(G_1), \mathbb P(G_2)\right) \leq W_1\left(\mathbb P(G_1), \mathbb P(G_2)\right) \leq \sqrt{\mathrm{MMD}^2\left(\mathbb P(G_1), \mathbb P(G_2)\right)+\tau}.
\]
\twocolumn

\end{document}